%% file: main.tex
\pdfoutput=1

\documentclass{article}

\usepackage[table,dvipsnames,svgnames]{xcolor}

\usepackage{assets/iclr2025, times}
\iclrfinalcopy

\usepackage{subcaption}
\usepackage[T1]{fontenc}
\usepackage[utf8]{inputenc}
\usepackage[nopatch=footnote]{microtype}
\usepackage{inconsolata}
\usepackage[scaled=0.85]{helvet}
\usepackage{booktabs}       
\usepackage{nicefrac}       
\usepackage{url}            
\usepackage[colorlinks=true,allcolors=blue]{hyperref}
\usepackage{fnpct}           
\usepackage{tcolorbox}       
\usepackage{amsfonts}
\usepackage{amssymb}
\usepackage{mathtools}
\usepackage{amsmath}
\usepackage{scalerel}
\usepackage{csquotes}        
\usepackage{xspace}          
\makeatletter
\xspaceaddexceptions{\csq@qclose@i}
\makeatletter
\usepackage{multirow}
\usepackage{cleveref}
\crefname{section}{\S}{\S\S}
\Crefname{section}{\S}{\S\S}
\crefformat{section}{\S#2#1#3}
\crefname{figure}{Fig.}{Fig.}
\crefname{appendix}{App.}{}
\crefname{equation}{eq.}{eqs.}
\crefname{table}{Table}{Tables}

\usepackage{tabularray}
\usepackage{cellspace}
\newcommand\cincludegraphics[2][]{\raisebox{-0.2\height}{\includegraphics[#1]{#2}}}

\usepackage[table-column-type=s]{siunitx}
\newcommand{\q}[2]{\qty[mode=math]{#1}{#2}\xspace}
\DeclareSIUnit[quantity-product = {}, reset-math-version = false]\thousand{k}
\DeclareSIUnit[quantity-product = {}, reset-math-version = false]\million{M}
\DeclareSIUnit[quantity-product = {}, reset-math-version = false]\billion{B}
\DeclareSIUnit[quantity-product = {}, reset-math-version = false]\trillion{T}
\DeclareSIUnit[quantity-product = {}, reset-math-version = false]\x{x}
\DeclareSIUnit[quantity-product = {}, reset-math-version = false]\percent{\%}
\DeclareSIUnit[quantity-product = {}, reset-math-version = false]\hour{h}
\DeclareSIUnit[quantity-product = {}, reset-math-version = false]\min{m}
\DeclareSIUnit[quantity-product = {}, reset-math-version = false]\sec{s}

\newcommand{\integer}[1]{%
    \num[
        mode = math,
        round-mode=places,
        round-precision=0,
        group-separator={,},
        group-minimum-digits=4,
        ]{#1}%
    \xspace
}

\newcommand{\float}[2][1]{%
    \num[
        group-digits=false, 
        round-precision=#1, 
        round-mode=places,
    ]{#2}%
    \xspace}

\xspaceaddexceptions{\textsubscript}

\makeatletter
\DeclareRobustCommand*{\escapeus}[1]{%
  \begingroup\@activeus\scantokens{#1\endinput}\endgroup}
\begingroup\lccode`\~=`\_\relax
   \lowercase{\endgroup\def\@activeus{\catcode`\_=\active \let~\_}}
\makeatother

\newcommand{\myemph}[1]{\textsf{{\escapeus{#1}}}}

\newtcolorbox{summary}{
  colframe=gray!75!black, 
  boxrule=0pt, 
  leftrule=2pt, 
  rightrule=2pt, 
  toprule=.5pt,
  bottomrule=.5pt,
  boxsep=4pt, 
  left=0pt, 
  right=0pt, 
  top=0pt, 
  bottom=0pt, 
  arc=2pt, 
  rounded corners=north,
}

\usepackage[textsize=tiny]{todonotes}

\newcommand{\defn}[1]{\textbf{#1}}  

\definecolor{14M}{HTML}{003049}
\definecolor{31M}{HTML}{2a9d8f}
\definecolor{70M}{HTML}{d62828}
\definecolor{160M}{HTML}{9b5de5}
\definecolor{410M}{HTML}{ff6b00} 

\usepackage{ifthen}
\newcommand{\pythia}[1]{%
    \ifthenelse{\equal{#1}{14M}}{\myemph{\textcolor{14M}{\q{14}{\million}}}}{%
    \ifthenelse{\equal{#1}{31M}}{\myemph{\textcolor{31M}{\q{31}{\million}}}}{%
    \ifthenelse{\equal{#1}{70M}}{\myemph{\textcolor{70M}{\q{70}{\million}}}}{%
    \ifthenelse{\equal{#1}{160M}}{\myemph{\textcolor{160M}{\q{160}{\million}}}}{%
    \ifthenelse{\equal{#1}{410M}}{\myemph{\textcolor{410M}{\q{410}{\million}}}}{%
    \PackageError{pythia}{The string `#1` is not in the mapping. Available options are 14M, 31M, 70M, 160M, or 410M}{Add the string to the mapping in the \pythia command.}%
    }}}}}%
    \xspace
}

\DeclareMathOperator*{\expect}{\mathbb{E}}
\newcommand{\kl}{\mathrm{KL}}
\newcommand{\loss}{\mathcal{L}}
\newcommand{\lossmdl}{\loss_{\myemph{MDL}}}
\newcommand{\lossce}{\loss_{\myemph{CE}}}
\newcommand{\x}{x}
\newcommand{\y}{y}
\newcommand{\xvec}{\boldsymbol{{x}}}
\newcommand{\yvec}{\boldsymbol{y}}
\newcommand{\weight}{\alpha}
\newcommand{\weightvec}{\boldsymbol{\alpha}}
\newcommand{\probe}{\boldsymbol{\theta}}
\newcommand{\latent}{\repr}
\newcommand{\repr}{\boldsymbol{\mathrm{h}}}
\newcommand{\dataset}{\mathcal{D}}
\newcommand{\R}{\mathbb{R}}
\newcommand{\modeldim}{d}

\newcommand{\pos}{\myemph{PoS}\xspace}
\newcommand{\dep}{\myemph{Dep}\xspace}
\newcommand{\semtag}{\myemph{SemTag}\xspace}
\newcommand{\coref}{\myemph{Coref}\xspace}
\newcommand{\ner}{\myemph{NER}\xspace}
\newcommand{\senti}{\myemph{Senti}\xspace}
\newcommand{\topic}{\myemph{Topic}\xspace}

\newcommand{\size}[1]{{\vert #1 \vert}}
\newcommand{\smalldots}{...}

\newcommand{\blimpgender}{\myemph{BLiMP (Gender Agreement)}\xspace}
\newcommand{\crowspairs}{\myemph{CrowS-Pairs}\xspace}
\newcommand{\crowspairsgender}{\myemph{CrowS-Pairs (Gender)}\xspace}
\newcommand{\simplebias}{\myemph{Simple Co-occurrence Bias}\xspace}

\newcommand{\arceasy}{\myemph{ARC (Easy)}\xspace}
\newcommand{\arcchallenge}{\myemph{ARC (Challenge)}\xspace}
\newcommand{\lambada}{\myemph{LAMBADA}\xspace}
\newcommand{\piqa}{\myemph{Piqa}\xspace}
\newcommand{\winogrande}{\myemph{WinoGrande}\xspace}
\newcommand{\wsc}{\myemph{WSC}\xspace}
\newcommand{\logiqa}{\myemph{Logiqa}\xspace}
\newcommand{\sciq}{\myemph{SciQ}\xspace}

\newcommand{\markquestion}[1]{\textit{\enquote{#1}}}

\definecolor{state0}{HTML}{377eb8}
\definecolor{state1}{HTML}{ff7f00}
\definecolor{state2}{HTML}{4daf4a}
\definecolor{state3}{HTML}{f781bf}
\definecolor{state4}{HTML}{a65628}

\DeclareRobustCommand{\circled}[2][]{%
    \tikz[baseline=(char.base)]{%
        \node[shape=circle,draw,inner sep=1pt, fill=#1] (char) {\normalfont{\small #2}};%
    }%
}
\newcommand{\statezero}{\circled[state0]{\myemph{0}}}
\newcommand{\stateone}{\circled[state1]{\myemph{1}}}
\newcommand{\statetwo}{\circled[state2]{\myemph{2}}}
\newcommand{\statethree}{\circled[state3]{\myemph{3}}}
\newcommand{\statefour}{\circled[state4]{\myemph{4}}}

\newcommand{\modelsizes}{\mathcal{M}}

\definecolor{lightgrey}{gray}{0.85}

\title{PolyPythias: Stability and Outliers across Fifty Language Model Pre-Training Runs}

\author{
  \!Oskar van der Wal\thanks{Equal contribution. 
  \footnotemark[2]\;\,Senior author. 
  \footnotemark[3]\;\,Work done while at EleutherAI. 
  Correspondence to: Stella Biderman <\href{mailto:stella@eleuther.ai}{\myemph{stella@eleuther.ai}}>, Oskar van der Wal <\href{mailto:oskar.vanderwal@gmail.com}{\myemph{oskar.vanderwal@gmail.com}}>, and Pietro Lesci <\href{mailto:pl487@cam.ac.uk}{\myemph{pl487@cam.ac.uk}}>.
  }\,\,\footnotemark[3]
  \\University of Amsterdam
  \And
  Pietro Lesci\footnotemark[1]\\University of Cambridge
  \And
  Max M\"{u}ller-Eberstein\\IT University of Copenhagen
  \AND
  Naomi Saphra\\Harvard University
  \And
  Hailey Schoelkopf\footnotemark[3]\\Anthropic
  \And
  Willem Zuidema\\University of Amsterdam
  \And
  Stella Biderman\footnotemark[2]\\EleutherAI
}

\begin{document}
\maketitle

\input{body}

\end{document}

%% file: body.tex
\vspace{-6pt}
\begin{abstract}
The stability of language model pre-training and its effects on downstream performance are still understudied. 
Prior work shows that the training process can yield significantly different results in response to slight variations in initial conditions, e.g., the random seed.
Crucially, the research community still lacks sufficient resources and tools to systematically investigate pre-training stability, particularly for decoder-only language models.
We introduce the PolyPythias, a set of \integer{45} new training runs for the Pythia model suite: \integer{9} new seeds across \integer{5} model sizes, from \q{14}{\million} to \q{410}{\million} parameters, resulting in about \q{7}{\thousand} new checkpoints that we release.
Using these new \integer{45} training runs, in addition to the \integer{5} already available, we study the effects of different initial conditions determined by the seed---i.e., parameters' initialisation and data order---on (i) downstream performance, (ii) learned linguistic representations, and (iii) emergence of training phases.
In addition to common scaling behaviours, our analyses generally reveal highly consistent training dynamics across both model sizes and initial conditions.
Further, the new seeds for each model allow us to identify outlier training runs and delineate their characteristics.
Our findings show the potential of using these methods to predict training stability.

\begin{tblr}{colspec = {X[c,m]  X[c,m]}, stretch = 1}
    \cincludegraphics[width=1.2em, keepaspectratio]{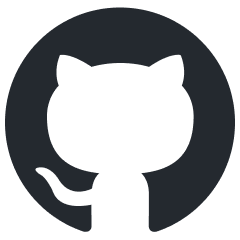} {\small \href{https://github.com/EleutherAI/pythia}{\myemph{EleutherAI/pythia}}}
    & \cincludegraphics[width=1.1em, keepaspectratio]{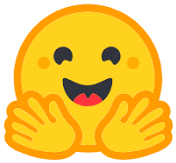} {\small\href{https://huggingface.co/collections/EleutherAI/polypythias-67bed6916110c8933e1ea561}{\myemph{EleutherAI/PolyPythias}}}
\end{tblr}
\end{abstract}

\vspace{-7pt}
\section{Introduction}

Training deep learning models, including contemporary large-scale transformer-based language models (LMs), is an inherently stochastic process in which \defn{randomness factors} \citep{pecher-etal-2024-survey}, such as the initialisation of model parameters and data shuffling, play a crucial role \citep{pham2020problems,gundersen2023sources}.
Prior work on LM adaptation (\citealp{dodge-etal-2020-finetuning, mccoy-etal-2020-berts, agarwal2021sensitivity, mosbach2021on}; \textit{inter alia}) and pre-training (\citealp{damour2022underspecification, madaan2024quantifying, alzahrani2024benchmarks}; \textit{inter alia})
shows that even slight variations in these randomness factors can lead to substantially different outcomes.
Specifically, training multiple times using the same implementation, hardware, dataset, and hyperparameters can, nonetheless, lead to large deviations in the final performance. 
This variability in performance has significant implications, primarily because conclusions drawn from single training runs may be misleading or incomplete. Thus, a systematic investigation into the stability of LM pre-training is essential to ensure robustness, reproducibility, and trustworthiness in applications that use these models \citep{sellam2022the, goldfarbtarrant2024multicontrievers}.
The stability of LM performance to randomness factors in their pre-training is still underexplored, especially for recent decoder-only architectures (e.g., \citealp{radford2019language}). 
Moreover, studying learning dynamics while ensuring coverage across these randomness factors is increasingly compute-intensive due to the size of contemporary LMs and datasets.

In this work, we define \defn{stability} as the change in a metric of interest (e.g., validation loss) caused by changes in randomness factors and quantify it using the standard deviation of that metric (see \citealp{du-nguyen-2023-measuring} for other approaches to quantify stability).
To provide a basis for analysing the stability of LMs to randomness factors (e.g., their training dynamics or final performance) without incurring the costs to train contemporary LMs, we introduce the \defn{PolyPythias}: an extension of the Pythia model suite \citep{biderman-etal-2023-pythia} trained on the Pile dataset \citep{gao-etal-2020-pile}, for which we release \integer{9} new training runs for \integer{5} model sizes, from \q{14}{\million} up to \q{410}{\million} parameters.
These new \integer{45} training runs---in addition to the \integer{5} already available in the suite---cover approximately \q{7}{\thousand} checkpoints across pre-training, and enable us to analyse training stability of large-scale transformer-based LM with respect to model size, parameter initialisation, and data order as quantified by metrics along the entire model training pipeline: downstream performance and consistency of predictions (\cref{sec:downstream_perf}), shifts in linguistic representations (\cref{sec:linguistic_probing}), and dynamics of the model parameters and training phases (\cref{sec:training_maps}).

By studying the PolyPythias, we find that: (i) language modelling is largely stable and follows predictable scaling laws with respect to downstream performance; (ii) across training, we identify consistent learning phases: an initial learning phase between steps $10^3$--$10^4$ and a critical learning phase between steps $10^4$--$10^5$; (iii) using training maps constructed from statistics of the model parameters, we identify the characteristics of stable training runs and the early signals of instability. 

In the following sections, we describe the PolyPythias release (\cref{sec:release_description}) and how we use the multiple training runs per model size to study the stability of models across various stages of the model training pipeline (\cref{sec:downstream_perf}--\cref{sec:training_maps}).
We conclude by combining the insights from the individual analyses (\cref{sec:discussion}).

\section{Extending the Pythia Suite: PolyPythias Release Description}
\label{sec:release_description}

The Pythia model suite \citep{biderman-etal-2023-pythia}---with its open data and weights for multiple model sizes, intermediate checkpoints, and detailed reporting of the training configurations---allows researchers to study the learning dynamics of realistic LMs without the need to train them from scratch.
Since its release, the suite has been extensively used to study, e.g., LMs' learning dynamics \citep{michaelov-bergen-2023-emergent, arnold2024phase}, memorisation patterns \citep{biderman2024emergent, lesci-etal-2024-causal}, and biases \citep{hu2024generative}.
The suite is composed of \integer{10} models ranging in size from \q{14}{\million} to \q{12}{\billion} parameters trained on the Pile dataset \citep{gao-etal-2020-pile, biderman-etal-2022-datasheet}, a \q{300}{\billion}-token curated collection of English documents.\footnote{There exists a deduplicated version of the Pile dataset used to train a second version of the Pythia suite.} 
All models are trained using the same data. Specifically, the dataset is shuffled and \enquote{packed} into sequences of \integer{2049}\footnote{Target tokens are the right-shifted input tokens; thus, an additional token is required to achieve the desired input and target sequence length of \integer{2048} tokens.} tokens. 
Training was performed using a cosine learning rate schedule with warm-up, and using a batch size of \integer{1024} sequences, resulting in exactly \q{143}{\thousand} optimisation steps.
In the original Pythia suite, for each model size, a single training run is available and consists of \integer{154} checkpoints: at initialisation (step \integer{0}), log-spaced up to step \q{1}{\thousand} (steps $1, 2, \smalldots, 512$), and every \q{1}{\thousand} steps afterwards (steps \q{1}{\thousand}--\q{143}{\thousand}).

We consider models with \pythia{14M}, \pythia{31M}, \pythia{70M}, \pythia{160M}, and \pythia{410M} parameters.
For each size, we release \integer{9} additional training runs resulting in about \q{7}{\thousand} new checkpoints. In \cref{app:release}, \cref{tab:pythia-models-links}, we report the links to the model checkpoints.
Each training run uses the same hyperparameters, codebase, and data as \citet{biderman-etal-2023-pythia} but varies the seeds for parameter initialisation and batch composition\footnote{We will use the term \enquote{batch composition} instead of \enquote{data order} because the GPT-NeoX codebase \citep{gpt-neox-library}, used to train the (Poly)Pythias, shuffles documents before packing them into sequences. This results in sequences that are not simply reshuffled across seeds; they are unique due to the different packing.}.
We use the standard (i.e., non-deduplicated) version of the Pile and release the tokenised and pre-shuffled datasets corresponding to the different seeds. 
More training details are in \cref{app:training_details}.

A limitation of our suite is that it spans model sizes up to \q{410}{\million} parameters. This choice reflects computational constraints, prioritising seed exploration and checkpoint granularity over scaling up model size. Our aim is to provide an additional resource for researchers unable to train even \q{410}{\million} parameter models from scratch, thus enabling them to study training stability across model sizes. 

Prior work that released multi-seed model suites includes \citet{sellam2022the} who introduced the MultiBERTs, a set of \integer{25} BERT-base (final) checkpoints trained with similar hyper-parameters to the original encoder-only BERT architecture \citep{devlin-etal-2019-bert} but with different random seeds. However, only \integer{28} intermediate checkpoints are available for \integer{5} of the runs, and the release is limited to encoder-only models in a single size. 
\citet{karamcheti2021mistral} introduced \integer{10} GPT-2 (\q{124}{\million} and \q{355}{\million} parameters; \citealp{radford2019language}) training runs, each with \integer{600} intermediate checkpoints, but still limited to two model sizes.
More recently, \citet{madaan2024quantifying} trained a suite of \integer{10} Llama-2-7B \citep{touvron2023llama2} models initialised with different random seeds on \q{210}{\billion} tokens, analysing \integer{21} intermediate checkpoints which, however, remain publicly unavailable.
PolyPythia compares favourably to these suites by spanning \integer{5} model sizes with \integer{154} checkpoints per run, using \integer{10} seeds per model, for a total of almost \q{7}{\thousand} checkpoints trained on publicly available data.

\section{Stability of Downstream Performance}
\label{sec:downstream_perf}

\begin{summary}
    We start by asking a key question, especially relevant for practitioners: \markquestion{How stable is model performance on downstream tasks to randomness factors?}
    We first study how performance varies across seeds (controlling both batch composition and parameters' initialisation) for a given model size. Then, we analyse how models' predictions and learned gender biases vary throughout the training process and across seeds.
    We find that language modelling is largely stable and follows predictable scaling laws with respect to downstream performance.
\end{summary}

Following \citet{biderman-etal-2023-pythia}, we measure model performance as the average \defn{accuracy} on a set of multiple-choice tasks: \arceasy and \arcchallenge \citep{clark2018arc}, \lambada \citep{paperno-etal-2016-lambada}, \logiqa \citep{logiqa}, \piqa \citep{bisk-piqa}, \sciq \citep{welbl-etal-2017-crowdsourcing}, \winogrande \citep{sakaguchi-etal-wino}, and \wsc \citep{wsc}.
We measure how predictions agree across seeds and throughout training by computing the Cohen's $\kappa$ \citep{Cohen1960} on the individual multiple-choice answers, where $\kappa\mathop{=}1$ means perfect agreement while $\kappa\mathop{=}0$ denotes agreement at the chance level.
Specifically, we define the \defn{inter-seed agreement} as the Cohen's $\kappa$ between the predictions of a model trained with a particular seed and the same model trained with seed \integer{0}.
Additionally, we define \defn{self-consistency} as the Cohen's $\kappa$ between the predictions of a model at the last checkpoint and any previous one.
Finally, we measure a model's gender bias as its accuracy on the \blimpgender \citep{warstadt-etal-2020-blimp-benchmark}, \crowspairsgender \citep{nangia-etal-2020-crows}, and \simplebias \citep{smith2022using} benchmarks.
More details about these benchmarks in \cref{app:experimental_details}.

\begin{figure}[!t]
    \centering
    \includegraphics[width=\linewidth]{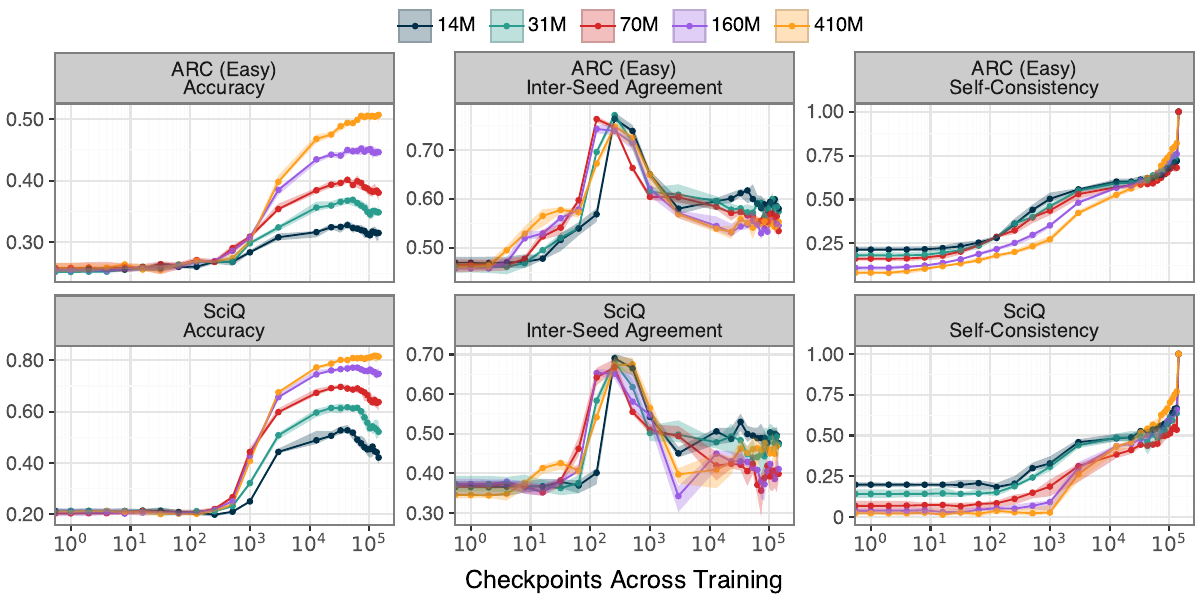}
    \vspace{-18pt}
    \caption{Accuracy, Inter-Seed Agreement, and Self-Consistency (median and interquartile range across seeds) on \arceasy and \sciq.}%
    \label{fig:downstream_main}
    \vspace{-10pt}
\end{figure}

\begin{figure}[!t]
    \centering
    \includegraphics[width=\linewidth]{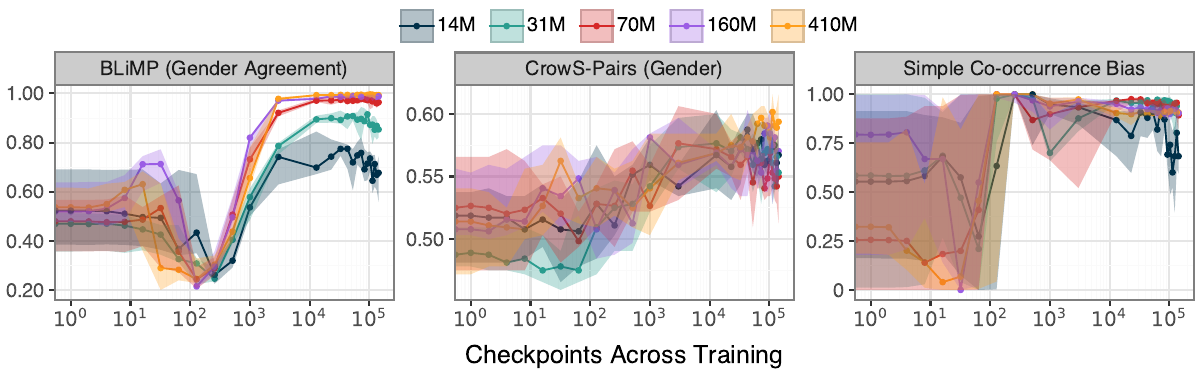}
    \vspace{-18pt}
    \caption{Accuracy, proportions of times the stereotypical answer is chosen, and proportion of times the male option is preferred for, respectively, \blimpgender, \crowspairsgender, \simplebias (median and interquartile range across seeds).}
    \label{fig:gender_bias}
    \vspace{-4pt}
\end{figure}

We perform the evaluation using the \myemph{Language Model Evaluation Harness} framework\footnote{\href{https://github.com/EleutherAI/lm-evaluation-harness}{\myemph{github.com/EleutherAI/lm-evaluation-harness}}.} \citep{eval-harness, biderman-etal-2024-evalharness}.
To limit computational costs while being able to track model behaviour across training for each size and seed, we evaluate performance on a subset of the available checkpoints. Specifically, we use checkpoints at (log-spaced) steps $0, 1, 2, \smalldots, 512$, \q{1}{\thousand}, and from step \q{3}{\thousand} onwards we choose every \q{10}{\thousand}-th step up to \q{143}{\thousand}, the final checkpoint.
We report accuracy, inter-seed agreement, and self-consistency on \arceasy and \sciq in \cref{fig:downstream_main} and the gender bias results in \cref{fig:gender_bias}. We show the other benchmarks in \cref{app:downstream_perf}.
For each metric, we show the median and interquartile range across seeds. We discuss the individual results below.

\paragraph{Downstream Performance.}
We find (unsurprisingly) that the larger models consistently outperform their smaller counterparts, as indicated by higher accuracy (\cref{fig:downstream_main}, left column), except for the more challenging tasks for which all models perform as good as random (see \arcchallenge, \logiqa, \wsc, \winogrande in \cref{app:downstream_perf}, \cref{fig:downstream_all}), a result consistent with \citealp{biderman-etal-2023-pythia}.
Performance improves most after step $10^3$. However, it drops between step $10^4$ and $10^5$ for all but \pythia{410M}, especially for the smallest models.
This finding aligns with prior work showing that smaller LMs suffer from \enquote{saturation}, i.e., a drop in performance at a later stage of the training process due to a mismatch between the dimension of the model representations and the high rank of the output embedding matrix \citep{michaelov-bergen-2023-emergent, godey2024small}.

\paragraph{Inter-Seed Agreement and Self-Consistency.}
For a more fine-grained analysis of how much model behaviour changes across seeds, we compare inter-seed agreement and self-consistency in \cref{fig:downstream_main} (middle and right column, respectively). 
Inter-seed agreement peaks at $\kappa \mathop{\approx} 0.7$ around step $10^3$ before converging towards a \enquote{moderate agreement}---i.e., $\kappa \mathop{\approx} 0.5$---around step $10^4$ and remaining at this level until the end of training.
Self-consistency steadily increases to \enquote{moderate agreement}---i.e., $\kappa \mathop{\approx} 0.5$---up to step $10^3$, after which it plateaus up to step $10^4$, before continuing to increase until the end of training as the models settle on their final answers. This finding aligns with the slow convergence observed in small models by \citet{diehl-martinez-etal-2024-tending}.

\paragraph{What is special about step $\mathbf{10^3}$?}
A qualitative analysis of model predictions, across tasks and seeds, shows that around step $10^3$ models begin generating non-random answers and accuracy improves.
Since PolyPythias use a batch size of slightly over \q{2}{\million} tokens, this step aligns with the \q{2.5}{\billion}-token mark identified by \citet{olsson2022context} as the point where induction heads form and in-context learning capabilities emerge. 
Also, this aligns with \citet{tigges2024llm} who observe a variety of circuits emerge in Pythia models, regardless of model size, between \q{2}{\billion} and \q{10}{\billion} tokens (i.e., steps \q{1}{\thousand}–\q{5}{\thousand})---e.g., induction heads, successor heads \citep{gould2024successor}, copy-suppression heads \citep{mcdougall2023copy}, and name-mover heads \citep{wang2022interpretability}.
Further, \citet{chang-bergen-2022-word} find that early in training LMs primarily rely on unigram token frequencies before gradually shifting to more contextual predictions---a finding further corroborated by \citet{meister-etal-2023-natural}. Similarly, \citet{jumelet2024black} show that Pythia models develop adjective order preferences within this same training range.
Collectively, these findings suggest that core semantic functions emerge at a consistent stage in training, regardless of model size or randomness factors.

\paragraph{Finding outlier seeds based on Accuracy and Validation Loss.}
Zooming in on \cref{fig:downstream_main}, we see that per-size accuracy is generally consistent across seeds, with two exceptions.
For a given model size, we can identify outlier seeds as follows.\footnote{A formal statistical test (e.g., ANOVA and Tukey's test) would have required a larger sample size (i.e., more tasks). Nonetheless, this simple heuristic allows us to discover the same outliers we find using other approaches.}
First, we consider the accuracy of the last checkpoints on the \arceasy, \lambada, \piqa, and \sciq tasks, for which all models perform better than random.
Second, for each task and model size, we standardise accuracy to have mean zero and a standard deviation of one by standardising across seeds.
Finally, we define a region of \integer{2} standard deviations from the mean of a model on that task and consider \enquote{outliers} those model-seed combinations that fall outside this region.
We (remarkably) find only two such combinations: \pythia{410M} seed \integer{3} and \integer{4}; only for these very seeds we observe \enquote{loss spikes} (see \cref{fig:train_loss} in \cref{app:training_details}).
We will further explore these outlier seeds using other metrics and approaches in \cref{sec:training_maps}. 

\paragraph{Gender Bias.}
We find distinct phases in the development of gender, both grammatical and bias, in \cref{fig:gender_bias}. Specifically, the models start to learn gender agreement between step $10^2$ and $10^3$ (see \blimpgender), which coincides with a sudden shift to a strong bias for using \enquote{male identifier words}  (see \simplebias).
Around this step, but less sharply, we also see an increase in gender bias measured by the \crowspairsgender benchmark, which measures more semantically diverse stereotypes than simple co-occurrence statistics.
We posit that the large variance observed for the bias measures reflects the poor reliability (e.g., due to small benchmark size, poor quality test items, etc) rather than actual bias differences (see \citealp{van2024undesirable, delobelle-etal-2024-metrics}).

\section{Representational Stability of Linguistic Information}
\label{sec:linguistic_probing}

\begin{summary}
    We now focus on the step before output generation and analyse token representations and find that they remain similar across seeds. Also, representational stability follows consistent trajectories across model sizes, suggesting that trends in smaller models reliably predict those in larger ones. 
\end{summary}

We study representational (in)stability by applying the information-theoretic probing approach proposed in \citet{muller-eberstein-etal-2023-subspace}.\footnote{For space reasons, we only sketch the method here and refer to the original paper for a detailed introduction.}
We consider seven linguistically motivated token-classification tasks: coreference resolution (\coref; \citealp{pradhan-etal-2013-towards}), dependency parsing (\dep; \citealp{silveira-etal-2014-gold}), named entity recognition (\ner; \citealp{pradhan-etal-2013-towards}), part-of-speech tagging (\pos; \citealp{pradhan-etal-2013-towards}), semantic tagging (\semtag; \citealp{abzianidze-etal-2017-parallel}), sentiment analysis (\senti; \citealp{socher-etal-2013-recursive}), and topic classification (\topic; \citealp{lang95news}); more details in \cref{app:experimental_details}.
For each task, we train a \defn{probe} classifier $\probe\in\R^{\modeldim\times\size{\mathcal{Y}}}$ as follows.
First, given an input-output pair, $(\xvec, \yvec) \mathop{\in} \dataset$, we collect the model representations, $\repr_l(\x)\mathop{\in}\R^\modeldim$, for each token $\x$ in the input sequence $\xvec$ at each layer $l$. 
Then, we aggregate per-layer representations into a global representation for that token using a learned weighting scheme $\weightvec\in\R^L$, obtaining $\latent(\x) \mathop{=} \sum_{l=1}^{L} \weight_l\, \repr_l(\x)$. 
This global representation is finally passed as input to the probe, which outputs class probabilities for that token.
Both $\weightvec$ and $\probe$ are jointly learned by optimising the minimum description length (MDL) loss:\footnote{We refer to \citet{voita-titov-2020-information} for a formal derivation.}
\begin{align}
    \lossmdl
    = - \expect_{\probe \sim p(\probe)} \Biggl[\sum_{(\xvec,\yvec)\in\dataset} 
    \underbrace{\Biggl(\sum_{\x \in\xvec, \y\in\yvec} \overbrace{\lossce\left(\probe^\top \latent(\x), y\right)}^{\text{Token-level loss}}\Biggr)}_{\text{Sequence-level loss}} 
    \Biggr]
    + \kl\big(p(\probe) \,\Vert\, q(\probe)\big)
    \label{eq:mdl}
\end{align}
where $\lossce$ is the cross-entropy loss. 
Probes trained with only cross-entropy loss can correctly map random representations to labels \citep{voita-titov-2020-information, pimentel-etal-2020-information}. Thus, \cref{eq:mdl} includes a KL-divergence term to keep the probe’s distribution $p(\probe)$ close to a sparsity-inducing prior $q(\probe)$.

We study representational stability through three metrics.
First, we measure the \defn{information content} of representations using the probe's macro-F1 score.
Second, we note that \cref{eq:mdl} corresponds to the probe's codelength; thus, we measure \defn{representational efficiency} by defining the \defn{codelength ratio} between random\footnote{We use the probes obtained from the randomly initialized models at step \integer{0}.} \textit{vs.}\ probe's representations where values close to \integer{0} indicate high representational efficiency.
Finally, we measure the \defn{representational shift} between two consecutive checkpoints using the principal \defn{subspace angles} (\defn{SSA}s; \citealp{knyazev2002ssa}): given two probes, SSAs return an angle between $\integer{0}^{\circ}$ and $\integer{90}^{\circ}$ where lower values represent more similar representations.

\begin{figure}[!t]
    \centering
    \includegraphics[width=\linewidth]{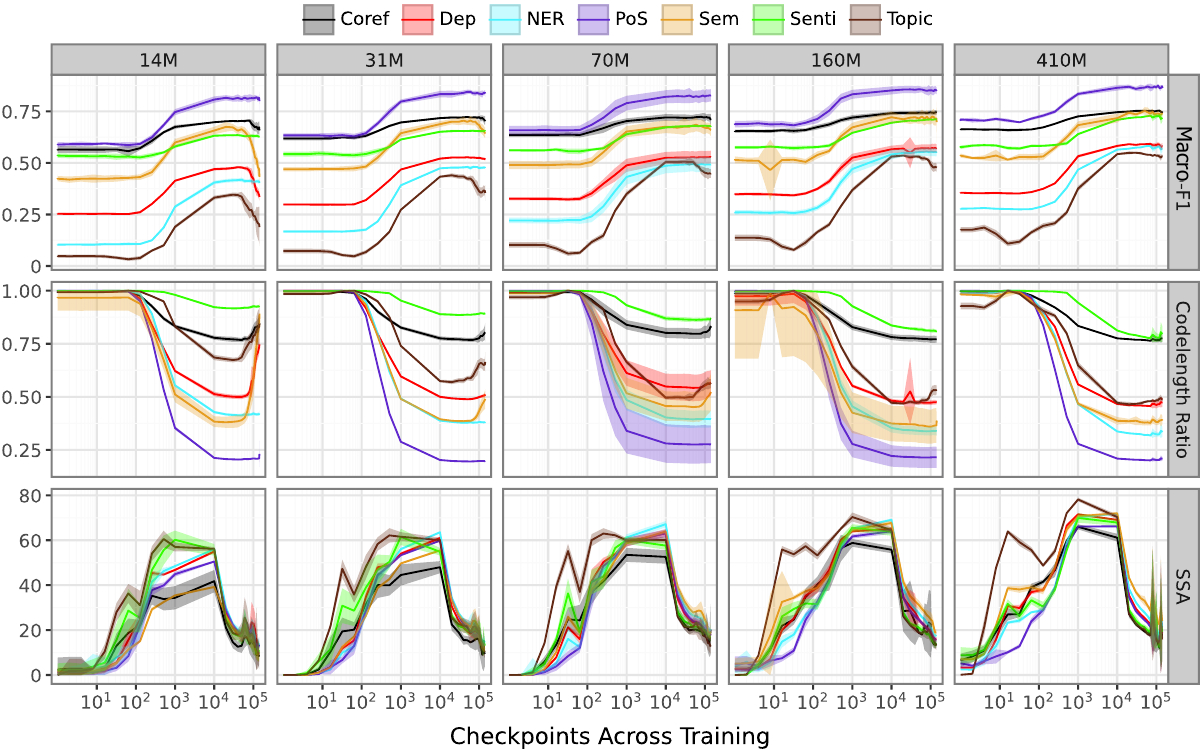}
    \vspace{-18pt}
    \caption{Changes in latent representations of linguistic phenomena (coreference, syntactic dependencies, named entities, parts-of-speech, semantic tags, sentiment, topic). 
    macro-F1, Codelength Ratio, and SSA (rows) for each linguistic task (colour) averaged across seeds (line) and the respective standard deviation (shaded area) for each model size (columns) and checkpoint (x-axis).}
    \label{fig:information-theoretic-probing}
    \vspace{-4pt}
\end{figure}

Results are summarised in \cref{fig:information-theoretic-probing}, which shows the macro-F1, Codelength Ratio, and SSAs (rows) for each task (colour) averaged across seeds (line) and the respective standard deviation (shaded area) for each model size (columns) and checkpoint (x-axis).
We discuss these results and explore their correlation across different model sizes below.

\paragraph{Information Content.} 
The macro-F1 scores mirror (unsurprisingly) the scaling behaviour observed for the \arceasy and \sciq tasks (\cref{sec:downstream_perf}), where scores consistently improve with model size across both lower-level syntactic tasks (e.g., \pos, \dep) and higher-level semantic ones (e.g., \ner, \senti).
In other words, larger models return representations with higher information content. 
While final values differ across model sizes, their trajectory is consistent: mirroring accuracy scores in \cref{fig:downstream_main} (top row), we observe initial improvements around step $10^3$ and a rapid increase in macro-F1 up to step $10^4$, after which performance for most tasks remains stable, except for the smaller model sizes.

\paragraph{Representational Efficiency.} 
Like macro-F1, the Codelength Ratio starts improving around step $10^3$ and reaches its stable value around step $10^4$.
The final Codelength Ratio for each task depends on the model size, with a distinction between models below and above \q{100}{\million} parameters. Specifically, we observe similar final values for syntactic tasks (e.g., \pos, \dep, \coref) across model sizes, while larger models achieve a \integer{5}-\q{10}{\percent} abs.\ lower Codelength Ratio on semantic tasks (e.g., \ner, \senti).

\paragraph{Representational Shift.} 
The SSAs follow (remarkably) similar trajectories across model sizes. For each task and model size, SSA increases up to step $10^3$, then plateaus until step $10^4$, and finally decreases until the end of training to a value around $\integer{20}^\circ$.
In other words, model representations after step $10^4$ tend to become more stable, matching the reduced rate of change in macro-F1 and Codelength Ratio.
The SSA does not directly correlate with macro-F1 or Codelength Ratio: larger changes in SSA early on have positive effects, while smaller changes later on can negatively affect these metrics (e.g., \pythia{14M}).
However, we note that SSA roughly follows the learning rate schedule: peaking around step \q{2}{\thousand} (end of the warm-up phase) and then decreasing until the end of training.

\paragraph{Correlation across model sizes.} 
We investigate the similarity of the trajectories of representational stability metrics across model sizes.
A metric's trajectory for a model size is based on the concatenation of its values for all tasks across all checkpoints. We first compute the Pearson correlation $r_{i,j}$ between a metric's trajectory for each pair of model sizes $\langle i,j\rangle \mathop{\in} \modelsizes\mathop{\times}\modelsizes$ where $\modelsizes = \{\pythia{14M}, \pythia{31M}, \pythia{70M}, \pythia{160M}, \pythia{410M}\}$.
To obtain the average $\overline{r}$ across all pairs of model sizes, we use the Fisher transformation over correlation coefficients \citep{fisher1970statistical}:
$\overline{r} \mathop{=} \mathrm{tanh}\left(\frac{1}{\size{\modelsizes}^2}\sum_{i,j \in \modelsizes\mathop{\times}\modelsizes} \mathrm{arctanh}\left(r_{i, j}\right)\right)$.
Finally, we average the resulting $\overline{r}$'s across seeds in the same way to obtain per-metric average correlations.
We find that macro-F1, Codelength Ratio, and SSA are highly correlated across model sizes with an average $\overline{r}$ of \float[2]{0.99} for macro-F1, \float[2]{0.98} for Codelength Ratio, and \float[2]{0.94} for SSA (all with a \textit{p}-value smaller than \float[3]{0.001}). This indicates that the representational stability of smaller models is strongly indicative of that of larger models.

\section{Training Phases and Outlier Seeds}
\label{sec:training_maps}

\begin{summary}
    After analysing the stability of performance and intermediate representations, we now examine the dynamics of model parameters using training maps. We find mostly consistent training dynamics across model sizes and seeds, with some exceptions. Additionally, training maps from smaller models can predict those of larger models and their final performance.
\end{summary}

We investigate the dynamics of model parameters using training maps. A \defn{training map} \citep{hu2023latent} associates each checkpoint with a latent state by fitting a Hidden Markov Model (HMM; \citealp{baum1966statistical}) to a vector of statistics derived from the model parameters (e.g., $L_2$-norm).
To fit the HMM, we first gather statistics---listed in \cref{app:training_maps}, \cref{tab:hmm_metrics}---from all checkpoints.
For each checkpoint within each model size, we standardise the statistics across seeds so that they have a mean of zero and a standard deviation of one, as HMMs are sensitive to the scale of the inputs. 
We use the standardised sequence to train the HMM with the Baum-Welch algorithm \citep{baum1970maximization}. 
Typically, the number of latent states, the primary hyperparameter in HMMs, is determined by minimising some information criterion \citep[e.g.,][]{schwarz1978estimating, akaike1998information} computed on a validation set.
To enable comparisons across model sizes, we use \integer{5} states, as this value is near optimal for the Bayesian Information Criterion across all sizes.
Consequently, for each training run, we obtain a sequence of latent states (one per each checkpoint) representing its training map.

\begin{figure}[!t]
    \centering
    \includegraphics[width=0.925\linewidth]{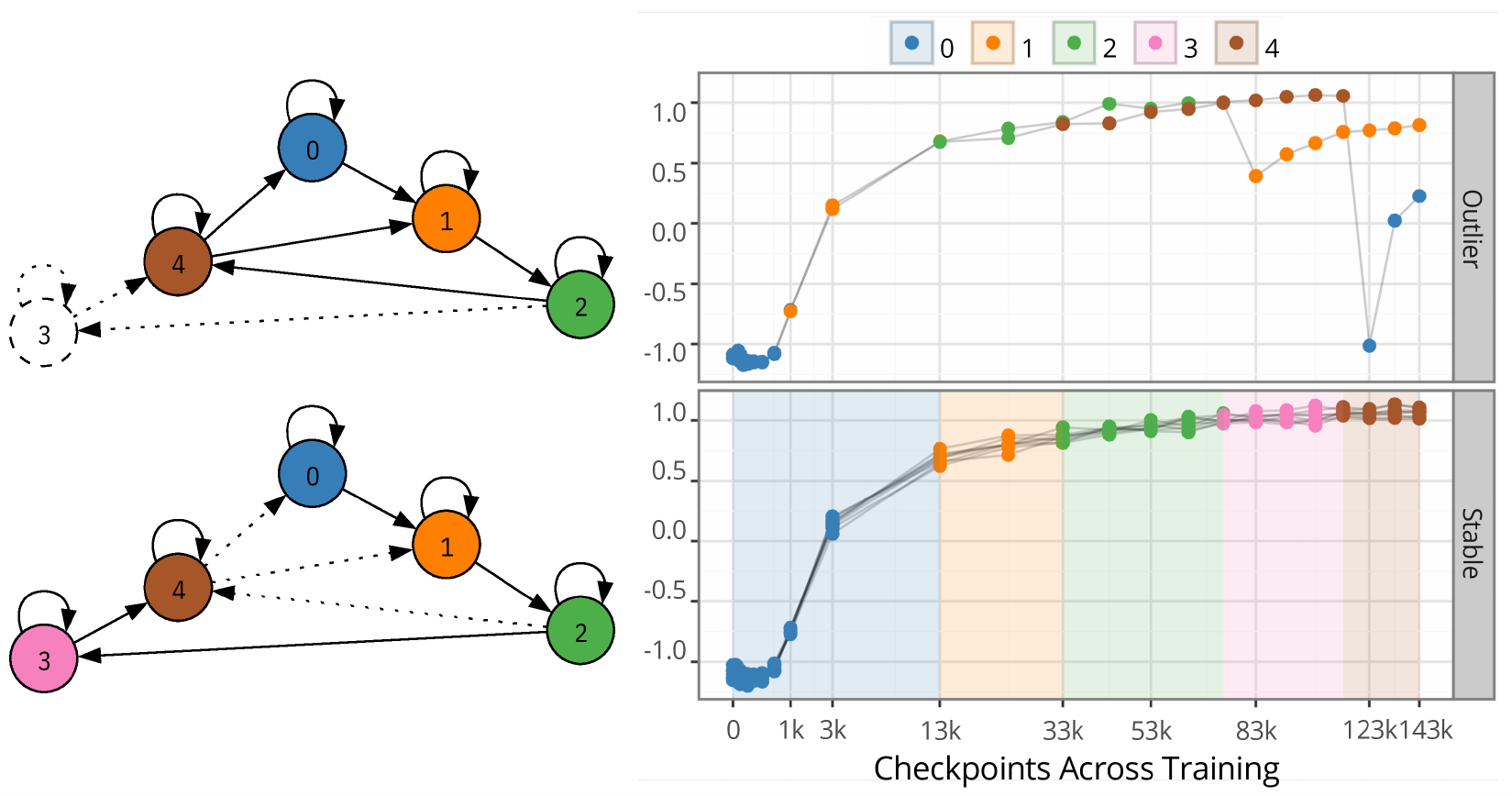}
    \vspace{-8pt}
    \caption{Average standardised accuracy (z-score) across the datasets \arceasy, \lambada, \piqa, and \sciq (right) for each seed of the model size \protect\pythia{410M}, along with their corresponding training maps (left). The HMM transitions are colour-coded according to the training map. The results are divided into outlier runs (top; seeds \integer{3} and \integer{4}) and stable runs (bottom).}
    \label{fig:phases_410m_outliers}
\end{figure}

First, we use training maps to compare training runs and find outliers (\cref{fig:phases_410m_outliers}), and we report which properties of the parameters drive state transitions (\cref{tab:outliers}).
Second, we study the relationship between training maps and final model performance (\cref{tab:overview_training_maps,tab:OLS_preds_410}), and investigate whether it is possible to zero-shot predict final performance from the training map alone. Specifically, we represent each training map as a \defn{bag-of-states}, i.e., a vector counting the number of times a model visits each state. We then use this as input for a linear regression model to predict final performance.
As our performance metric, we use the average accuracy of the final model checkpoint on \arceasy, \lambada, \piqa, and \sciq; we choose these tasks as all models perform better than random. 
To average accuracy across tasks consistently, for each model size, we standardise it across seeds to have a mean of zero and a standard deviation of one. We refer to the resulting scores as \textbf{z-scores}.

\paragraph{Characterising the training maps of outlier seeds.}
Across model sizes and seeds, we find that training maps are linear graphs
with a few exceptions for specific seeds, i.e.,  seed \integer{3} and \integer{4} of model size \pythia{410M}.
In \cref{fig:phases_410m_outliers}, we visualise the HMM (left) and training map (right) for \pythia{410M} and show outlier (top) and stable (bottom) seeds separately.
In line with \citet{hu2023latent}, we find that linear training maps (\cref{fig:phases_410m_outliers}, bottom-left) describe stable dynamics and performance (\cref{fig:phases_410m_outliers}, bottom-right), while maps with \enquote{forks}---i.e., regressions to an earlier state---are associated with instability. Specifically, forks are only present in the training maps of the outlier seeds (\cref{fig:phases_410m_outliers}, top-left); these seeds showcase sudden drops in performance (\cref{fig:phases_410m_outliers}, top-right) and loss spikes (see \cref{fig:train_loss} in \cref{app:training_details}).

\paragraph{Drivers of state transition.}
In \cref{tab:outliers}, we report the three main drivers of state transition, focusing on the transitions that appear in the stable maps but not in the outlier maps and those that are only present in the outlier maps.
First, outlier maps fail to perform the transition $\statetwo\rightarrow\statethree$ in which the parameters $L_2$, median bias, and the average weight variance decrease.
Abnormal state transitions for the outlier maps ($\statetwo\rightarrow\statefour$) are driven by an increase in the variance of the weights' singular values ($\sigma_\lambda$).
The subsequent state transitions ($\statefour\rightarrow\statezero$ and $\statefour\rightarrow\stateone$) coincide with a strong performance drop and are driven by a sharp decrease in $\sigma_\lambda$.
This phenomenon is also described by \citet{godey2024small} as \enquote{representation degeneration}: the distribution of singular values first becomes increasingly uniform and then abruptly degenerates around a point.
We leave further exploration as future work. 

\begin{table}[!t]
    \centering
    \small{\input{assets/tables/transition_features}}
    \vspace{-6pt}
    \caption{Top three drivers of state transitions for \protect\pythia{410M}. Focus on transitions unique to stable maps and those only present in outlier maps.}
    \label{tab:outliers}
    \vspace{-4pt}
\end{table}

\paragraph{Consistency in state transitions.}
In \cref{tab:overview_training_maps} (right), we report the steps at which state transitions occur. We find that state transitions mostly happen around the same step for each model size and across different seeds. 
The only exception is \pythia{410M}, for which we observe variability caused by outlier seeds.
Additionally, transitions tend to occur at similar steps: the first transition happens around step $10^3$, followed by two more transitions between $10^3$ and $10^4$, and the final transition around $10^5$. Notably, \pythia{410M} is less consistent, even when we do not account for the outlier seeds.
We leave the exploration of this aspect for future work.

\paragraph{Training maps and final performance.}
The training maps (interestingly) reveal that the outlier runs start to deviate from the other runs long before they show worse scores on the performance metrics. In \cref{fig:phases_410m_outliers} (top), we see that both runs enter state $\statetwo$ prematurely and fail to make the transition to state $\statethree$.
To investigate whether a model's training maps are informative of that model's final performance, we perform linear regressions using the bag-of-states to predict the final average z-score. We report the regression $R^2$ (\cref{tab:overview_training_maps}, $R^2$ column). Only the training map of \pythia{410M}  is predictive of performance ($R^2\mathop{=}0.99$). Also, this model, due to the outlier seeds, is associated with a high performance variance (\cref{tab:overview_training_maps}, $\sigma^2$ column).
We report the regression coefficients for \pythia{410M} in the last row of \cref{tab:OLS_preds_410}. We observe that the outlier seeds (\integer{3} and \integer{4}) are the only ones receiving a negative coefficient. In other words, the bag-of-states obtained from the training map of \pythia{410M} has enough information to predict this model will underperform.

\paragraph{Predicting performance across model sizes using training maps.}
We investigate whether training maps from smaller models can help predict the performance of a larger model. Specifically, we attempt to predict the performance of \pythia{410M} using training maps from smaller models.
Our approach consists of two steps. First, for a given small model (e.g., \pythia{14M}), we train an HMM on checkpoints from all seeds. We then use this HMM to assign each checkpoint of \pythia{410M} to a latent state, effectively performing a \enquote{zero-shot} prediction of its training map. Second, we aggregate the training maps of the large model across all seeds into a bag-of-states representation, which we use to predict the performance of each seed. Our hypothesis is that if this regression successfully predicts model performance, then the training map of the smaller model carries meaningful information that can be used to predict the performance of the larger model.
In \cref{tab:OLS_preds_410}, we report the $R^2$ and coefficients of the regression that uses each size-specific training map to predict the performance of \pythia{410M}. We find that forming the bag-of-states using predict the average z-score successfully with $R^2\mathop{>}0.9$ for all models, except \pythia{14M}, the smallest size. Also, the negative coefficients for the outlier seeds indicate that it is possible to predict underperforming runs across sizes.
When computing the bag-of-states for \pythia{410M}, we pass all its checkpoints through the HMM to generate the full training map. Ideally, we would like to use only the initial checkpoints to predict the final performance of a partially trained model. This would allow us to decide early on whether to stop a specific run.
However, when we construct the bag-of-states using only a partial training run, we fail to predict the average z-score accurately. Empirically, we find that at least \q{120}{\thousand} steps (out of the total \q{143}{\thousand}) are required for a reliable prediction.
We leave the investigation of which properties can be predicted from the bag-of-states of early checkpoint metrics as future work.

\begin{table}[!t]
    \centering
    \small{\input{assets/tables/overview_training_maps}}
    \vspace{-5pt}
    \caption{
    Overview of statistics for the training maps found for the different sizes of Pythia models.
   $R^2$: the goodness of fit for the linear regression for the bag-of-states and the z-scores. $\sigma^2$: the variance of the average performance z-scores. 
    $^\ast$Averaged across seeds, except for outlier seeds \integer{3} and \integer{4} for \protect\pythia{410M} to remove non-linearities due to forks in the training maps.
    }
    \label{tab:overview_training_maps}
\end{table}

\begin{table}[!t]
    \centering
    \resizebox{0.88\linewidth}{!}{%
    \input{assets/tables/zero-shot}
    }
    \vspace{-5pt}
    \caption{Linear regression coefficients returned by regressing the average z-score of \protect\pythia{410M} on its bag-of-states obtained from the zero-shot training map constructed using the HMM trained on the model listed in column \enquote{HMM}. Outlier seeds are highlighted in \colorbox{lightgrey}{grey}.}
    \label{tab:OLS_preds_410}
    \vspace{-4pt}
\end{table}

\section{Discussion}
\label{sec:discussion}

Our experiments on downstream performance (\cref{sec:downstream_perf}), intermediate representations (\cref{sec:linguistic_probing}), and model parameters (\cref{sec:training_maps}) allow us to examine the stability of training and find outlier runs using different methods across the model training pipeline. In this section, we analyse commonalities across the resulting metrics to identify broader characteristics of LM pre-training dynamics.

\paragraph{Language modelling is largely stable.} 
Generally, we observe LM pre-training dynamics to follow consistent trajectories. Across seeds and model sizes, downstream performance and representational efficiency consistently increase during pre-training, and training maps are linear (except for the outlier seeds). 
Furthermore, model scaling laws seem to hold across seeds, not only for downstream performance but also for information content and representational efficiency of model representations.
Similarly, both at the performance and representational level, we observe the effect of \enquote{saturation} \citep{michaelov-bergen-2023-emergent,godey2024small} in smaller models.

\paragraph{Linguistic information is encoded in the initial learning phase ($\mathbf{10^3}$--$\mathbf{10^4}$ steps).}
Across all experiments, metrics start moving away from the initial random baseline around step $10^3$ (\q{2}{\billion} tokens circa) and reach their convergence level around step $10^4$ (\q{20}{\billion} tokens circa).
In this initial phase, representational shift peaks and linguistic information begins to be encoded into the models' latent representations. 
Through the lens of multiple metrics, we can analyse model behaviour in this phase in detail. Specifically, while the amount and the efficiency with which linguistic information is encoded in model representations have already increased substantially at step $10^3$, the model does not yet generate coherent outputs, as indicated by low performance on linguistic acceptability benchmarks like \blimpgender.
Simultaneously, self-consistency is low while inter-seed agreement is high, which we hypothesise to be an artefact of all models initially choosing an incorrect baseline answer.
In terms of training maps, this phase corresponds to the $\statezero\rightarrow\stateone$ transition, which occurs consistently in this initial training phase for all model sizes (except for the \pythia{410M} outliers). 

\paragraph{Most improvements happen in the \enquote{critical} learning phase ($\mathbf{10^4}$--$\mathbf{10^5}$ steps).} 
In the range of $10^3$ to $10^4$ steps, most learning occurs, as measured by all of our metrics. Performance increases the most, and linguistic information content and representational efficiency converge to close-to-final values. 
At the same time, the representational shift begins to decrease from its peak, showing how the information encoded in the model begins to stabilise. 
This is reflected in terms of performance by the simultaneous increase in self-consistency. The fact that inter-seed agreement decreases before remaining relatively constant until the end of training indicates that models settle on their final answers after this stage.
In the training maps, this phase corresponds to the $\stateone\rightarrow\statetwo$ and $\statetwo\rightarrow\statethree$ transitions that occur before step $10^5$. 
The fact that our outlier runs (\pythia{410M}, seeds \integer{3} and \integer{4}) exit state $\statetwo$ prematurely indicates that this phase is important for the LM's downstream performance.
Furthermore, we note that these learning phases are (remarkably) similar and occur at the same time during pre-training as for encoder-only models \citep{muller-eberstein-etal-2023-subspace}. Also, they follow similar trajectories as recurrent LSTM architectures \citep{saphra-lopez-2019-understanding}. 
We leave the exploration of which modelling or data decisions may result in these learning phases for future work.
Finally, we observe that all benchmarks still improve past the optimal token count (e.g., \q{8.2}{\billion} tokens, or around \q{4}{\thousand}--\q{5}{\thousand} steps) predicted by the Chinchilla scaling law \citep{hoffmann-etal-2022-training}.

\paragraph{Training maps describe outlier seeds.}
Using the downstream performance results and training loss (\cref{sec:downstream_perf}), and training maps (\cref{sec:training_maps}), we identify outlier seeds and explain how the model parameters change in the unstable pre-training regime.
Moreover, we find that it is possible to predict training maps in a \enquote{zero-shot} fashion (e.g., using HMM trained on smaller models to predict the training map of larger ones), suggesting that statistics of the parameters of smaller models are informative of their larger counterparts.
We propose using PolyPythias to investigate potential connections between the state transitions observed in training maps and the emergence of components or behaviours in circuit analyses, as discussed in \cref{sec:downstream_perf}. 
However, a limitation is that early-training state transitions may occur too consistently across seeds for each model size to provide meaningful insights into this relationship. 
To better understand whether connections exist between circuit formation (e.g., the development of induction heads or other circuit components in each seed) and state transitions, interventional studies of training dynamics may be necessary.
We leave the exploration of this aspect for future work.

\section{Conclusions}

In this work, we introduce PolyPythia, a multi-seed extension of the Pythia model suite \citep{biderman-etal-2023-pythia}, adding \integer{45} extra pre-training runs for a total of \integer{10} seeds across \integer{5} model sizes. This expanded resource is designed to facilitate research on training dynamics and model stability.  
Through our experiments, we demonstrate the usefulness of PolyPythias by analysing the stability of the language modelling pipeline. We examine downstream performance, intermediate representations, and parameter training dynamics across seeds and model scales. Our findings suggest that, with some exceptions, language modelling remains largely stable.  

We hope these additional seeds will support further research into the impact of randomness in model pre-training. 
This includes studying the robustness of different evaluation metrics \citep[e.g.,][]{sellam2022the,van2024undesirable,madaan2024quantifying}, identifying factors contributing to suboptimal training \citep[e.g.,][]{zoph2022st,zhai2023stabilizing,chowdhery2023palm,zeng2022glm,chung2024stable}, and improving the predictability of model performance across scales \citep[e.g.,][]{kaplan2020scaling,srivastava2023beyond}.
Further, the diverse training runs with different data orders enable further studies on memorisation \citep[e.g.,][]{biderman2024emergent,lesci-etal-2024-causal} or on the relationship of the data to the emergence of learned behaviours \citep{wal2022birth,biderman-etal-2023-pythia,jumelet2024black,belrose2024neural}.
Lastly, PolyPythia provides a valuable testbed for assessing benchmark evaluations' reliability and model interventions' effectiveness.

\section*{Acknowledgements}
This research was made possible through computational resources generously provided by StabilityAI. 
We express our gratitude to Michael Hu for his assistance with the implementation of the HMM training maps presented in \cref{sec:training_maps}.
We thank Andreas Vlachos, Tiago Pimentel, and Clara Meister for their insightful feedback on earlier drafts. 
Finally, we extend our thanks to Davide Lesci and Marco Lesci for proofreading the final version of the manuscript.

\textbf{Oskar van der Wal} initiated this project during his research internship at EleutherAI and gratefully acknowledges their support. His work is partially funded by the Dutch Research Council (NWO) through the project \enquote{The biased reality of online media} (406.DI.19.059). The views expressed in this work do not represent any undisclosed current or future affiliations.

\textbf{Pietro Lesci} received funding from the European Research Council (ERC) under the European Union’s Horizon 2020 Research and Innovation programme grant AVeriTeC (Grant agreement No. 865958).

\textbf{Max M\"{u}ller-Eberstein} thanks the IT University of Copenhagen's High-Performance Computing Cluster for supporting the representational stability experiments in \cref{sec:linguistic_probing}.

\textbf{Naomi Saphra}'s work was enabled in part by a gift from the Chan Zuckerberg Initiative Foundation to establish the Kempner Institute for the Study of Natural and Artificial Intelligence.

\textbf{Hailey Schoelkopf} thanks EleutherAI for the opportunity to conduct this work. 

\textbf{Willem Zuidema} is funded by the Institute for Logic, Language and Computation of the University of Amsterdam.

\bibliography{biblio.bib}
\bibliographystyle{assets/acl_natbib}
\clearpage

\appendix

\section{Training Details}
\label{app:training_details}

We used the \myemph{v1.0} version of the GPT-Neox codebase\footnote{\href{https://github.com/EleutherAI/gpt-neox/tree/v1.0}{\myemph{github.com/EleutherAI/gpt-neox}} tag \myemph{v1.0}.} for model training.
We report the training loss for the two outlier pre-training runs, \pythia{410M} seed \integer{3} and \integer{4}, below. We refer to the \href{https://wandb.ai/eleutherai/pythia-extra-seeds}{\myemph{Weights \& Biases space}} for the training losses and logs of all runs.

\begin{figure}[!h]
    \centering
    \begin{subfigure}[t]{\linewidth}
        \centering
        \includegraphics[width=0.8\linewidth]{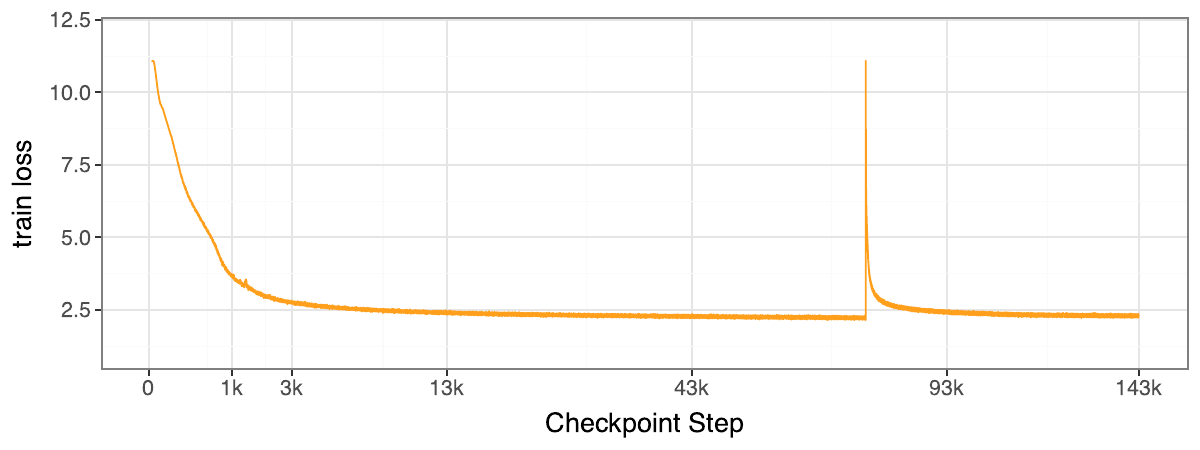}
        \caption{Seed \integer{3}}
    \end{subfigure}
    \begin{subfigure}[t]{\linewidth}
        \centering
        \includegraphics[width=0.8\linewidth]{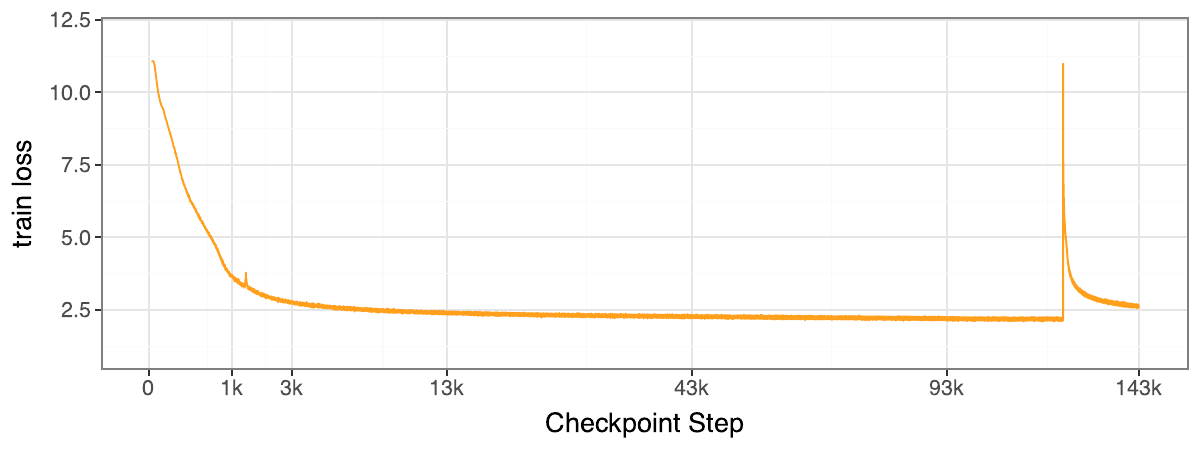}
        \caption{Seed \integer{4}}
    \end{subfigure}
    \caption{Training loss for \protect\pythia{410M} outlier seeds \integer{3} and \integer{4} showing \enquote{loss spikes}.}
    \label{fig:train_loss}
\end{figure}

\section{Evaluation Datasets Details}
\label{app:experimental_details}

In this section, we describe the datasets used in \cref{sec:downstream_perf} and \cref{sec:linguistic_probing}.

\paragraph{\arceasy and \arcchallenge \citep{clark2018arc}.}
The AI2 Reasoning Challenge (\myemph{ARC}) is a benchmark for evaluating natural science question answering. It consists of two subsets, Easy and Challenge, where questions are drawn from standardized science exams. Following \citet{brown-etal-2020-language}, evaluation is performed using log-likelihood-based scoring and reported in terms of accuracy.

\paragraph{\lambada \citep{paperno-etal-2016-lambada}.}
\lambada is a language modelling benchmark that measures a model’s ability to predict the final word of a sentence given its full context. The dataset consists of narrative texts, and successful prediction requires long-range context comprehension.

\paragraph{\logiqa \citep{logiqa}.}
\logiqa is a reading comprehension dataset designed to test logical reasoning in language models. The questions are derived from LSAT exam questions and require deductive reasoning and critical thinking skills. Evaluation is performed using multiple-choice accuracy.

\paragraph{\piqa \citep{bisk-piqa}.}
\piqa is a dataset designed to evaluate physical commonsense reasoning. It consists of multiple-choice questions requiring an understanding of everyday physical interactions. Evaluation is based on accuracy.

\paragraph{\sciq \citep{welbl-etal-2017-crowdsourcing}.}
\sciq is a question-answering dataset focused on scientific topics. It includes multiple-choice, direct-answer, and \enquote{cloze}-style questions sourced from educational materials. Evaluation involves measuring accuracy on multiple-choice and direct-answer formats.

\paragraph{\wsc \citep{wsc}.}
The Winograd Schema Challenge (WSC) is a coreference resolution task designed to test commonsense reasoning. It consists of sentence pairs that differ by a single word, requiring the model to correctly resolve ambiguous pronouns based on contextual cues. Accuracy is used as the primary evaluation metric.

\paragraph{\winogrande \citep{sakaguchi-etal-wino}.}
\winogrande is an expanded version of the Winograd Schema Challenge, containing sentence pairs with minor lexical variations. The task evaluates a model’s ability to resolve ambiguous pronouns by comparing the probabilities of different completions. Accuracy is reported as the primary metric.

\paragraph{\blimpgender \citep{warstadt-etal-2020-blimp-benchmark}.}
\blimpgender is a subset of the \myemph{BLiMP} benchmark designed to evaluate gender bias in language models. It consists of minimal sentence pairs that differ only in gender-marked words, allowing the assessment of whether a model exhibits gender preference in syntactic and morphological structures.

\paragraph{\crowspairs \citep{nangia-etal-2020-crows}.}
\crowspairs assesses biases in language models by presenting minimal sentence pairs that contrast stereotypical and non-stereotypical perspectives on US-protected demographic groups. We use the adaptation by \citet{neveol-etal-2022-french}, which removes instances identified as problematic for validity following \citet{blodgett-etal-2021-stereotyping}.

\paragraph{\simplebias \citep{smith2022using}.}
\simplebias evaluates gender associations in language models by analysing the likelihood of gendered identifiers (e.g., \enquote{man}, \enquote{woman}) appearing in simple template-based prompts. Following \citet{brown-etal-2020-language, smith2022using}, we report directional bias in model predictions. The dataset is available at \href{https://huggingface.co/datasets/oskarvanderwal/simple-cooccurrence-bias}{\myemph{huggingface.co/datasets/oskarvanderwal/simple-cooccurrence-bias}}.

\paragraph{\coref \citep{pradhan-etal-2013-towards}.}
\coref is a dataset for coreference resolution, which involves linking mentions of the same entity in a text. We use the coreference annotation layer from OntoNotes 5.0, which is commonly used to evaluate models' ability to resolve pronouns and named entity references in complex passages.

\paragraph{\dep \citep{silveira-etal-2014-gold}.}
\dep is a dataset for dependency parsing, providing annotations of relative syntactic functions based on the English Web Treebank. It is used to evaluate a model’s ability to predict grammatical relations between words in a sentence.

\paragraph{\ner \citep{pradhan-etal-2013-towards}.}
\ner (Named Entity Recognition) is a dataset used to identify and classify proper nouns into \integer{18} predefined categories, such as persons, organizations, and locations. We use the NER annotation layer from OntoNotes 5.0, which serves as a standard benchmark for evaluating entity recognition performance.

\paragraph{\pos \citep{pradhan-etal-2013-towards}.}
\pos (Part-of-Speech Tagging) is a dataset that annotates words in a sentence with one of \integer{51} syntactic categories (e.g., noun, verb, adjective), as taken from OntoNotes 5.0. It evaluates the models’ ability to perform syntactic parsing at the word level.

\paragraph{\semtag \citep{abzianidze-etal-2017-parallel}.}
\semtag is a dataset for semantic tagging, which classifies words or phrases based on \integer{69} semantic roles and meanings. It is used to assess the models' ability to understand and differentiate word meanings in context.

\paragraph{\senti \citep{socher-etal-2013-recursive}.}
\senti is a sentiment analysis dataset containing sentences labelled with sentiment polarity (positive, negative, neutral). It is used to evaluate models’ ability to infer sentiment from textual data.

\paragraph{\topic \citep{lang95news}.}
\topic is a dataset for topic classification, consisting of documents labelled with \integer{20} predefined topic categories. It serves as a benchmark for evaluating models’ ability to perform document-level topic classification.

\section{Training Maps}
\label{app:training_maps}

In \cref{tab:hmm_metrics}, we report the metrics used to fit the HMMs to create the training maps in \cref{sec:training_maps}. These metrics are (partially) taken from \citet{hu2023latent} Appendix B, to which we refer for further details.

\begin{table}[!ht]
    \centering
    \begin{tabular}{cl}
    \toprule
    \textbf{Metric} & \textbf{Description} \\
    \midrule
        $L_1$ & The $L_1$-norm, averaged over the weight matrices \\
        $L_2$ & The $L_2$-norm, averaged over the weight matrices \\
        $\nicefrac{L_1}{L_2}$ & Weight sparsity (ratio of their $L_1$ and $L_2$ norms), averaged over the weight matrices\\
        $\mu_w$ & Sample mean of the weights \\
        $\text{median}_w$ & Median of the weights \\
        $\sigma_w$ & Sample variance of weights \\
        $\mu_b$ & Sample mean of the biases \\
        $\text{median}_b$ & Median of the biases \\
        $\sigma_b$ & Sample variance of biases \\
        $\text{trace}$ & The average trace over the weight matrices \\
        $\lambda_{max}$ & The average spectral norm of the weights \\
        $\nicefrac{\text{trace}}{\lambda_{max}}$ & The average trace over spectral norm\\
        $\mu_\lambda$ & The average singular value over the weights \\
        $\sigma_\lambda$ & Sample variance of singular values over the weights \\
    \bottomrule
    \end{tabular}
    \caption{Statistics of model parameters (both weights and biases) used to fit the HMMs in \cref{sec:training_maps}.}
    \label{tab:hmm_metrics}
\end{table}

\section{Additional Figures}
\label{app:downstream_perf}

We report the figures for all the benchmarks analysed in \cref{sec:downstream_perf}. The figures follow in the next pages.

\begin{figure}[htbp]
    \centering
    \includegraphics[width=\linewidth]{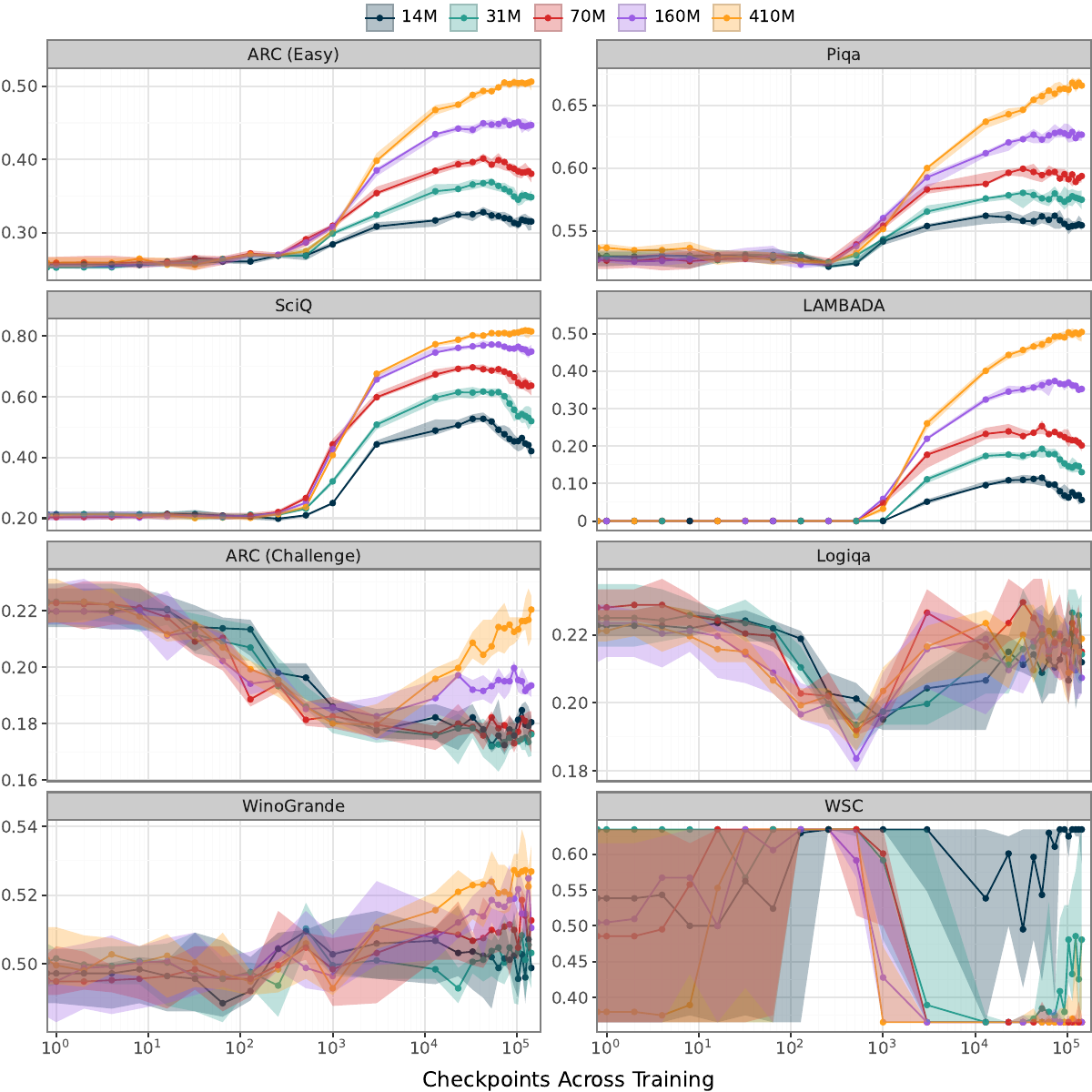}
    \vspace{-15pt}
    \caption{Accuracy (median and interquartile range across seeds) on \arceasy, \piqa, \sciq, \lambada, \arcchallenge, \logiqa, \winogrande, and \wsc tasks.}%
    \label{fig:downstream_all}
\end{figure}

\begin{figure}[htbp]
    \centering
    \includegraphics[width=\linewidth]{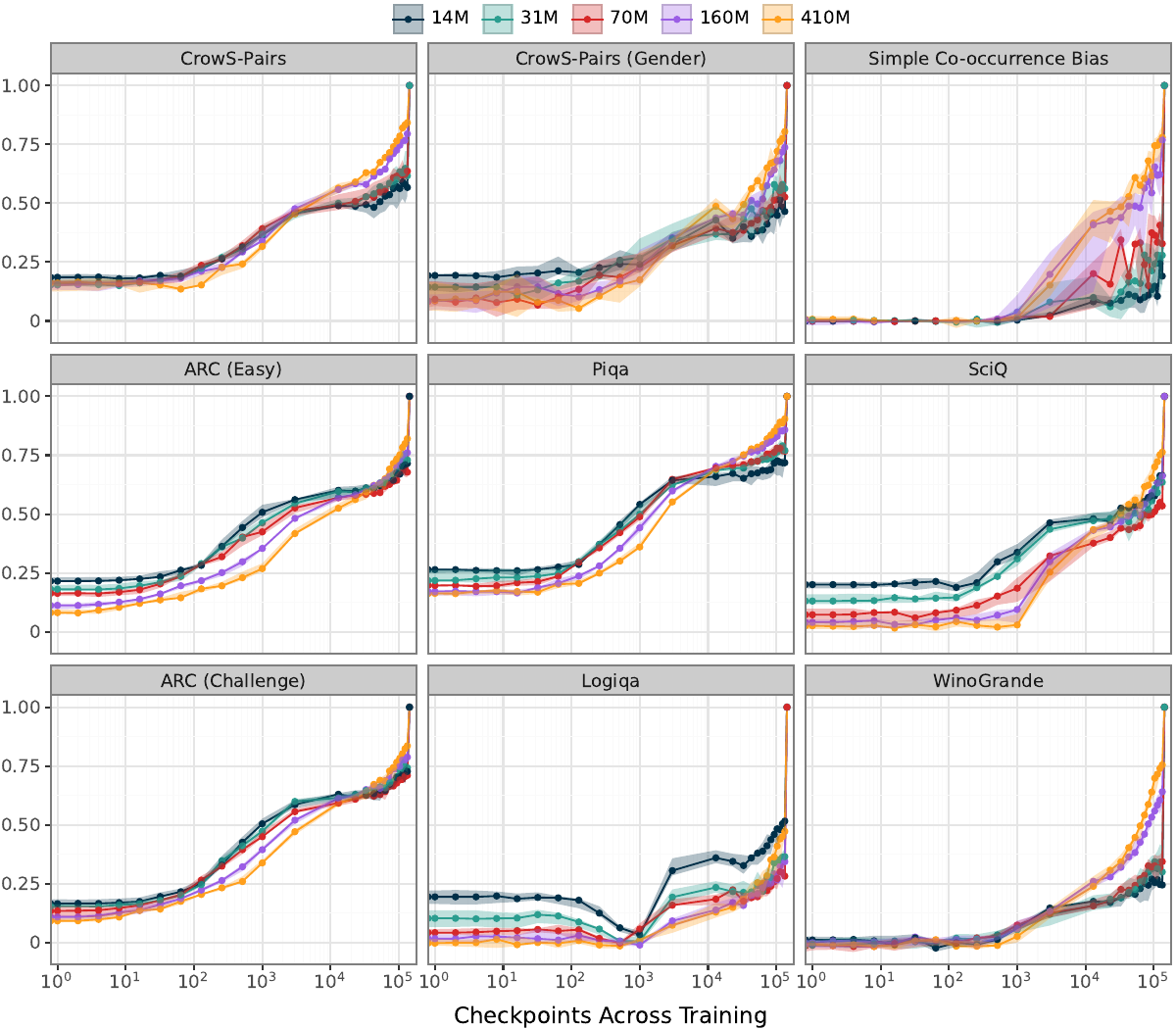}
    \vspace{-15pt}
    \caption{Self-consistency (median and interquartile range across seeds) on \crowspairs, \crowspairsgender, \simplebias, \arceasy, \arcchallenge, \piqa, \sciq, \arcchallenge, \logiqa, and \winogrande.}%
    \label{fig:self_consistency}
\end{figure}

\begin{figure}[htbp]
    \centering
    \includegraphics[width=\linewidth]{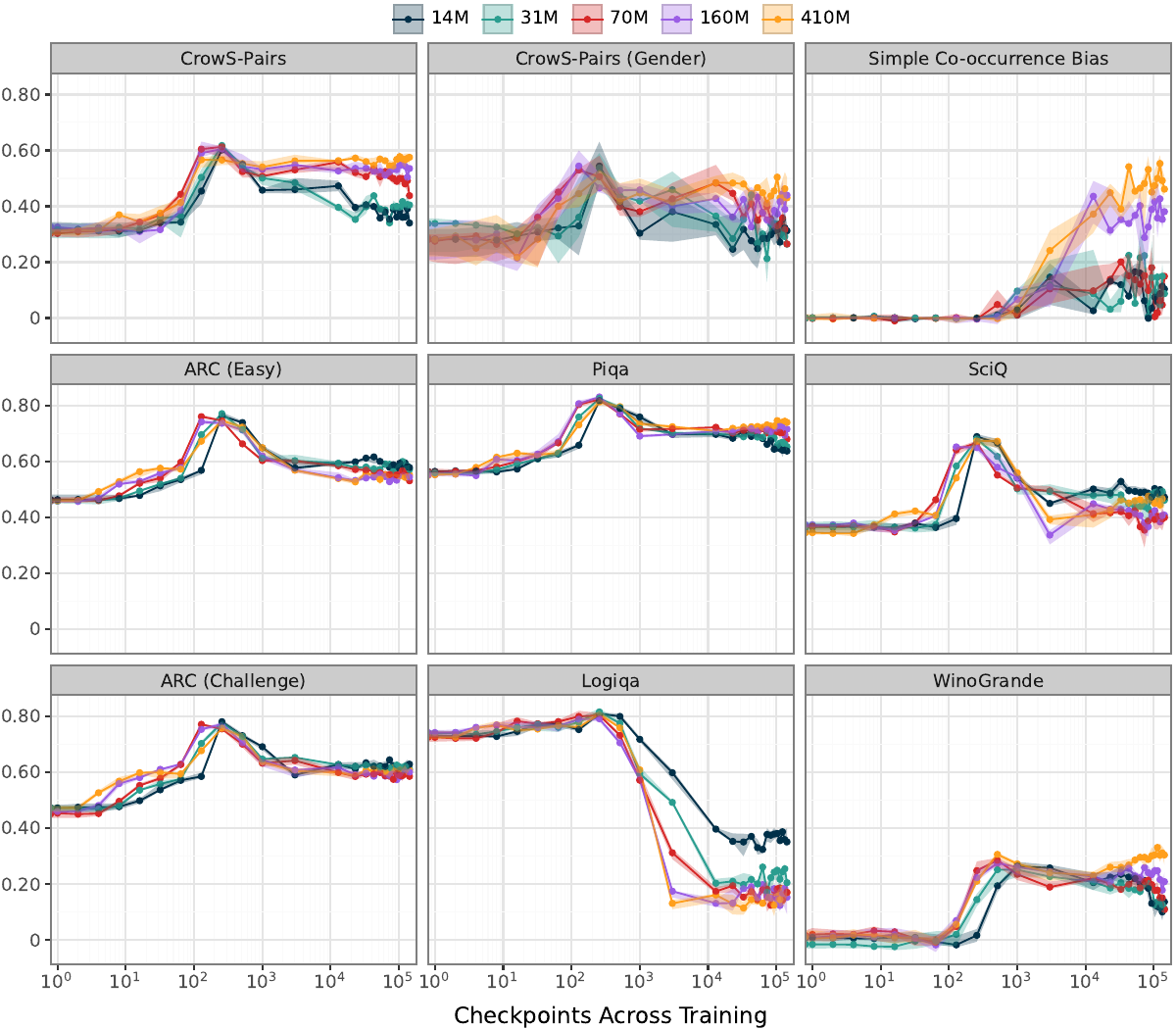}
    \vspace{-15pt}
    \caption{Inter-seed agreement (median and interquartile range across seeds) on \crowspairs, \crowspairsgender, \simplebias, \arceasy, \arcchallenge, \piqa, \sciq, \arcchallenge, \logiqa, and \winogrande.}%
    \label{fig:inter_agreement}
\end{figure}

\section{Release Links}
\label{app:release}

Besides the link to the Hugging Face collection listing all checkpoints reported in the abstract\footnote{Explicitly, \href{https://huggingface.co/collections/EleutherAI/polypythias-67bed6916110c8933e1ea561}{\myemph{huggingface.co/collections/EleutherAI/polypythias-67bed6916110c8933e1ea561}}.}, in \cref{tab:pythia-models-links} we report the links to the individual checkpoints on the Hugging Face Hub. Also, the indices used to recreate the pre-shuffled datasets are available at \href{https://huggingface.co/datasets/EleutherAI/pile-preshuffled-seeds}{\myemph{huggingface.co/datasets/EleutherAI/pile-preshuffled-seeds}}. Links follow on the next page (after the figures).

\begin{table}[!ht]
    \centering
    \small
    \input{assets/tables/model_links}
    \caption{Links to the individual checkpoints in the PolyPythias release.}
    \label{tab:pythia-models-links}
\end{table}

%% file: assets/tables/transition_features.tex
\newcommand{\maxeig}{\lambda_{\mathrm{max}}}

\begin{tabular}{crrrr}
    \toprule
    \textbf{Transition} & \textbf{Description} & \multicolumn{3}{c}{\textbf{Top \integer{3} feature determining each transition}} \\
    \midrule
    $\statetwo\rightarrow\statethree$ & Missing in Outlier & $\text{median}_b \downarrow \float[2]{0.44}$ & $L_2 \downarrow \float[2]{0.42}$ & $\sigma_w \downarrow \float[2]{0.5}$ \\
    \midrule
    $\statetwo\rightarrow\statefour$ & Only in Outlier & $\lambda_{\mathrm{max}} \uparrow \float[2]{1.41}$ & $\sigma_b \uparrow \float[2]{1.79}$ & $\sigma_\lambda \uparrow \float[2]{1.71}$ \\
    $\statefour\rightarrow\statezero$ & Only in Outlier & $L_1/L_2 \uparrow \float[2]{2.76}$ & $\sigma_b \downarrow \float[2]{2.43}$ & $\sigma_\lambda \downarrow \float[2]{2.99}$ \\
    $\statefour\rightarrow\stateone$ & Only in Outlier & $L_1/L_2 \uparrow \float[2]{2.2}$ & $\sigma_\lambda \downarrow \float[2]{2.21}$ & $\maxeig \downarrow \float[2]{1.94}$ \\
    \bottomrule
\end{tabular}

%% file: assets/tables/overview_training_maps.tex
\begin{tabular}{rcrrrrrr}
    \toprule
     & & \multicolumn{2}{c}{\textbf{z-scores}} & \multicolumn{4}{c}{\textbf{Step at which transition happens$^\ast$}} \\
      \cmidrule(lr){3-4} \cmidrule(lr){5-8}
    \textbf{Size} & 
    \textbf{Fork} & \textbf{$R^2$} &
    \textbf{$\sigma^2$} &
    $\statezero\rightarrow\stateone$ &
    $\stateone\rightarrow\statetwo$ &
    $\statetwo\rightarrow\statethree$ &
    $\statethree\rightarrow\statefour$
    \\
    \midrule
    \pythia{14M} &  & \float[2]{0.75} & \float[2]{0.651847} & $\q{5}{\thousand}_{\pm\q{0.7}{\thousand}}$ & $\q{38}{\thousand}_{\pm\q{3.5}{\thousand}}$ & $\q{72}{\thousand}_{\pm\q{2.2}{\thousand}}$ & $\q{108}{\thousand}_{\pm\q{2.7}{\thousand}}$ \\
    \pythia{31M} & & \float[2]{0.64} & \float[2]{0.844727} & $\q{7}{\thousand}_{\pm\q{0.7}{\thousand}}$ & $\q{42}{\thousand}_{\pm\q{2.6}{\thousand}}$ & $\q{69}{\thousand}_{\pm\q{2.1}{\thousand}}$ & $\q{104}{\thousand}_{\pm\q{2.0}{\thousand}}$ \\
    \pythia{70M}  & & \float[2]{0.41} & \float[2]{0.708686} & $\q{6}{\thousand}_{\pm\q{0.6}{\thousand}}$ & $\q{28}{\thousand}_{\pm\q{3.1}{\thousand}}$ & $\q{58}{\thousand}_{\pm\q{3.3}{\thousand}}$ & $\q{94}{\thousand}_{\pm\q{2.5}{\thousand}}$ \\
    \pythia{160M} & & \float[2]{0.03} & \float[2]{0.584236} & $\q{2}{\thousand}_{\pm 0}$ & $\q{18}{\thousand}_{\pm\q{0.8}{\thousand}}$ & $\q{61}{\thousand}_{\pm\q{1.5}{\thousand}}$ & $\q{100}{\thousand}_{\pm\q{1.6}{\thousand}}$ \\
    \pythia{410M} & \checkmark  & \float[2]{0.99} & \float[2]{0.976170} & $\q{18}{\thousand}_{\pm\q{22.6}{\thousand}}$ & $\q{35}{\thousand}_{\pm\q{24.1}{\thousand}}$ & $\q{73}{\thousand}_{\pm\q{20.9}{\thousand}}$ & $\q{114}{\thousand}_{\pm\q{4.9}{\thousand}}$ \\
    \bottomrule
\end{tabular}

%% file: assets/tables/zero-shot.tex
\begin{tabular}{rrrrr>{\columncolor{lightgrey}}r>{\columncolor{lightgrey}}rrrrrrr}
\toprule
& & \multicolumn{10}{c}{\textbf{Seed used to train} \pythia{410M}} \\
      \cmidrule(lr){3-12}
 \textbf{HMM} & $R^2$ & \integer{0} & \integer{1} & \integer{2} & \integer{3} & \integer{4} & \integer{5} & \integer{6} & \integer{7} & \integer{8} & \integer{9} \\
\midrule
\pythia{14M} & $\float[2]{0.17888706308024116}$ & $\float[2]{0.014195231503016714}$ & $\float[2]{0.05547733019971712}$ & $\float[2]{0.3935494189020172}$ & $\float[2]{-1.1210624826562245}$ & $\float[2]{-0.21285631132882799}$ & $\float[2]{0.034836280851370915}$ & $\float[2]{0.3519975025667783}$ & $\float[2]{0.014195231503016714}$ & $\float[2]{0.07611837954807044}$ & $\float[2]{0.3935494189020172}$ \\
\pythia{31M} & $\float[2]{0.9735187305636377}$ & $\float[2]{0.42161864807892435}$ & $\float[2]{0.5222846754597512}$ & $\float[2]{0.4834049670663356}$ & $\float[2]{-0.5591951228010664}$ & $\float[2]{-2.7847570215308446}$ & $\float[2]{0.42161864807892435}$ & $\float[2]{0.3055636346153987}$ & $\float[2]{0.32095262069809927}$ & $\float[2]{0.4415548432981211}$ & $\float[2]{0.4269541070363365}$ \\
\pythia{70M} & $\float[2]{0.9742258730960877}$ & $\float[2]{0.3392410877984066}$ & $\float[2]{0.4270726169254168}$ & $\float[2]{0.48308132687782884}$ & $\float[2]{-0.599772933406038}$ & $\float[2]{-2.751658963211933}$ & $\float[2]{0.2312758935351001}$ & $\float[2]{0.1946258566035364}$ & $\float[2]{0.5142680524668735}$ & $\float[2]{0.6733876694578028}$ & $\float[2]{0.4884793929529741}$ \\
\pythia{160M} & $\float[2]{0.9577765336386406}$ & $\float[2]{0.45609450349409286}$ & $\float[2]{0.45609450349409286}$ & $\float[2]{0.6926019693836558}$ & $\float[2]{-0.5826569720288315}$ & $\float[2]{-2.7361874968907984}$ & $\float[2]{0.31436865724166285}$ & $\float[2]{0.2753633803215183}$ & $\float[2]{0.3667428821237957}$ & $\float[2]{0.2533739705047098}$ & $\float[2]{0.5042046023560953}$ \\
\midrule
\pythia{410M} & $\float[2]{0.9861432660088311}$ & $\float[2]{0.39184002169299353}$ & $\float[2]{0.4468726382913197}$ & $\float[2]{0.3842079206084432}$ & 
$\float[2]{-0.566596948341907}$ & $\float[2]{-2.802573993309753}$ & $\float[2]{0.4709291798041819}$ & $\float[2]{0.47973403674970694}$ & $\float[2]{0.4158965632058558}$ & $\float[2]{0.49300519726857184}$ & $\float[2]{0.2866853840305753}$ \\
\bottomrule
\end{tabular}

%% file: assets/tables/model_links.tex
\begin{tabular}{lll}

\toprule

\textbf{Model Size} & \textbf{Seed} & \textbf{Links}\\

\midrule

\multirow{10}{*}{\pythia{14M}}
    & \integer{0} (default) & \href{https://huggingface.co/EleutherAI/pythia-14m}{\myemph{huggingface.co/EleutherAI/pythia-14m}} \\ 
    & \integer{1} & \href{https://huggingface.co/EleutherAI/pythia-14m-seed1}{\myemph{huggingface.co/EleutherAI/pythia-14m-seed1}} \\
    & \integer{2} & \href{https://huggingface.co/EleutherAI/pythia-14m-seed2}{\myemph{huggingface.co/EleutherAI/pythia-14m-seed2}} \\ 
    & \integer{3} & \href{https://huggingface.co/EleutherAI/pythia-14m-seed3}{\myemph{huggingface.co/EleutherAI/pythia-14m-seed3}} \\ 
    & \integer{4} & \href{https://huggingface.co/EleutherAI/pythia-14m-seed4}{\myemph{huggingface.co/EleutherAI/pythia-14m-seed4}} \\ 
    & \integer{5} & \href{https://huggingface.co/EleutherAI/pythia-14m-seed5}{\myemph{huggingface.co/EleutherAI/pythia-14m-seed5}} \\ 
    & \integer{6} & \href{https://huggingface.co/EleutherAI/pythia-14m-seed6}{\myemph{huggingface.co/EleutherAI/pythia-14m-seed6}} \\ 
    & \integer{7} & \href{https://huggingface.co/EleutherAI/pythia-14m-seed7}{\myemph{huggingface.co/EleutherAI/pythia-14m-seed7}} \\ 
    & \integer{8} & \href{https://huggingface.co/EleutherAI/pythia-14m-seed8}{\myemph{huggingface.co/EleutherAI/pythia-14m-seed8}} \\ 
    & \integer{9} & \href{https://huggingface.co/EleutherAI/pythia-14m-seed9}{\myemph{huggingface.co/EleutherAI/pythia-14m-seed9}} \\

\midrule

\multirow{10}{*}{\pythia{31M}}
    & \integer{0} (default) & \href{https://huggingface.co/EleutherAI/pythia-31m}{\myemph{huggingface.co/EleutherAI/pythia-31m}} \\
    & \integer{1} & \href{https://huggingface.co/EleutherAI/pythia-31m-seed1}{\myemph{huggingface.co/EleutherAI/pythia-31m-seed1}} \\
    & \integer{2} & \href{https://huggingface.co/EleutherAI/pythia-31m-seed2}{\myemph{huggingface.co/EleutherAI/pythia-31m-seed2}} \\
    & \integer{3} & \href{https://huggingface.co/EleutherAI/pythia-31m-seed3}{\myemph{huggingface.co/EleutherAI/pythia-31m-seed3}} \\
    & \integer{4} & \href{https://huggingface.co/EleutherAI/pythia-31m-seed4}{\myemph{huggingface.co/EleutherAI/pythia-31m-seed4}} \\
    & \integer{5} & \href{https://huggingface.co/EleutherAI/pythia-31m-seed5}{\myemph{huggingface.co/EleutherAI/pythia-31m-seed5}} \\
    & \integer{6} & \href{https://huggingface.co/EleutherAI/pythia-31m-seed6}{\myemph{huggingface.co/EleutherAI/pythia-31m-seed6}} \\
    & \integer{7} & \href{https://huggingface.co/EleutherAI/pythia-31m-seed7}{\myemph{huggingface.co/EleutherAI/pythia-31m-seed7}} \\
    & \integer{8} & \href{https://huggingface.co/EleutherAI/pythia-31m-seed8}{\myemph{huggingface.co/EleutherAI/pythia-31m-seed8}} \\
    & \integer{9} & \href{https://huggingface.co/EleutherAI/pythia-31m-seed9}{\myemph{huggingface.co/EleutherAI/pythia-31m-seed9}} \\

\midrule 

\multirow{10}{*}{\pythia{70M}}
    & \integer{0} (default) & \href{https://huggingface.co/EleutherAI/pythia-70m}{\myemph{huggingface.co/EleutherAI/pythia-70m}} \\
    & \integer{1} & \href{https://huggingface.co/EleutherAI/pythia-70m-seed1}{\myemph{huggingface.co/EleutherAI/pythia-70m-seed1}} \\
    & \integer{2} & \href{https://huggingface.co/EleutherAI/pythia-70m-seed2}{\myemph{huggingface.co/EleutherAI/pythia-70m-seed2}} \\
    & \integer{3} & \href{https://huggingface.co/EleutherAI/pythia-70m-seed3}{\myemph{huggingface.co/EleutherAI/pythia-70m-seed3}} \\
    & \integer{4} & \href{https://huggingface.co/EleutherAI/pythia-70m-seed4}{\myemph{huggingface.co/EleutherAI/pythia-70m-seed4}} \\
    & \integer{5} & \href{https://huggingface.co/EleutherAI/pythia-70m-seed5}{\myemph{huggingface.co/EleutherAI/pythia-70m-seed5}} \\
    & \integer{6} & \href{https://huggingface.co/EleutherAI/pythia-70m-seed6}{\myemph{huggingface.co/EleutherAI/pythia-70m-seed6}} \\
    & \integer{7} & \href{https://huggingface.co/EleutherAI/pythia-70m-seed7}{\myemph{huggingface.co/EleutherAI/pythia-70m-seed7}} \\
    & \integer{8} & \href{https://huggingface.co/EleutherAI/pythia-70m-seed8}{\myemph{huggingface.co/EleutherAI/pythia-70m-seed8}} \\
    & \integer{9} & \href{https://huggingface.co/EleutherAI/pythia-70m-seed9}{\myemph{huggingface.co/EleutherAI/pythia-70m-seed9}} \\

\midrule 

\multirow{16}{*}{\pythia{160M}}
    & \integer{0} (default) & \href{https://huggingface.co/EleutherAI/pythia-160m}{\myemph{huggingface.co/EleutherAI/pythia-160m}} \\
    & \integer{1} & \href{https://huggingface.co/EleutherAI/pythia-160m-seed1}{\myemph{huggingface.co/EleutherAI/pythia-160m-seed1}} \\
    & \integer{2} & \href{https://huggingface.co/EleutherAI/pythia-160m-seed2}{\myemph{huggingface.co/EleutherAI/pythia-160m-seed2}} \\
    & \integer{3} & \href{https://huggingface.co/EleutherAI/pythia-160m-seed3}{\myemph{huggingface.co/EleutherAI/pythia-160m-seed3}} \\
    & \integer{4} & \href{https://huggingface.co/EleutherAI/pythia-160m-seed4}{\myemph{huggingface.co/EleutherAI/pythia-160m-seed4}} \\
    & \integer{5} & \href{https://huggingface.co/EleutherAI/pythia-160m-seed5}{\myemph{huggingface.co/EleutherAI/pythia-160m-seed5}} \\
    & \integer{6} & \href{https://huggingface.co/EleutherAI/pythia-160m-seed6}{\myemph{huggingface.co/EleutherAI/pythia-160m-seed6}} \\
    & \integer{7} & \href{https://huggingface.co/EleutherAI/pythia-160m-seed7}{\myemph{huggingface.co/EleutherAI/pythia-160m-seed7}} \\
    & \integer{8} & \href{https://huggingface.co/EleutherAI/pythia-160m-seed8}{\myemph{huggingface.co/EleutherAI/pythia-160m-seed8}} \\
    & \integer{9} & \href{https://huggingface.co/EleutherAI/pythia-160m-seed9}{\myemph{huggingface.co/EleutherAI/pythia-160m-seed9}} \\
    
    \cmidrule{2-3}

    & \integer{1} (only data) & \href{https://huggingface.co/EleutherAI/pythia-160m-data-seed1}{\myemph{huggingface.co/EleutherAI/pythia-160m-data-seed1}} \\
    & \integer{2} (only data) & \href{https://huggingface.co/EleutherAI/pythia-160m-data-seed2}{\myemph{huggingface.co/EleutherAI/pythia-160m-data-seed2}} \\
    & \integer{3} (only data) & \href{https://huggingface.co/EleutherAI/pythia-160m-data-seed3}{\myemph{huggingface.co/EleutherAI/pythia-160m-data-seed3}} \\

    \cmidrule{2-3}
    
    & \integer{1} (only parameters) & \href{https://huggingface.co/EleutherAI/pythia-160m-weight-seed1}{\myemph{huggingface.co/EleutherAI/pythia-160m-weight-seed1}} \\
    & \integer{2} (only parameters) & \href{https://huggingface.co/EleutherAI/pythia-160m-weight-seed2}{\myemph{huggingface.co/EleutherAI/pythia-160m-weight-seed2}} \\
    & \integer{3} (only parameters) & \href{https://huggingface.co/EleutherAI/pythia-160m-weight-seed3}{\myemph{huggingface.co/EleutherAI/pythia-160m-weight-seed3}} \\

\midrule

\multirow{10}{*}{\pythia{410M}}
    & \integer{0} (default) & \href{https://huggingface.co/EleutherAI/pythia-410m}{\myemph{huggingface.co/EleutherAI/pythia-410m}} \\
    & \integer{1} & \href{https://huggingface.co/EleutherAI/pythia-410m-seed1}{\myemph{huggingface.co/EleutherAI/pythia-410m-seed1}} \\
    & \integer{2} & \href{https://huggingface.co/EleutherAI/pythia-410m-seed2}{\myemph{huggingface.co/EleutherAI/pythia-410m-seed2}} \\
    & \integer{3} & \href{https://huggingface.co/EleutherAI/pythia-410m-seed3}{\myemph{huggingface.co/EleutherAI/pythia-410m-seed3}} \\
    & \integer{4} & \href{https://huggingface.co/EleutherAI/pythia-410m-seed4}{\myemph{huggingface.co/EleutherAI/pythia-410m-seed4}} \\
    & \integer{5} & \href{https://huggingface.co/EleutherAI/pythia-410m-seed5}{\myemph{huggingface.co/EleutherAI/pythia-410m-seed5}} \\
    & \integer{6} & \href{https://huggingface.co/EleutherAI/pythia-410m-seed6}{\myemph{huggingface.co/EleutherAI/pythia-410m-seed6}} \\
    & \integer{7} & \href{https://huggingface.co/EleutherAI/pythia-410m-seed7}{\myemph{huggingface.co/EleutherAI/pythia-410m-seed7}} \\
    & \integer{8} & \href{https://huggingface.co/EleutherAI/pythia-410m-seed8}{\myemph{huggingface.co/EleutherAI/pythia-410m-seed8}} \\
    & \integer{9} & \href{https://huggingface.co/EleutherAI/pythia-410m-seed9}{\myemph{huggingface.co/EleutherAI/pythia-410m-seed9}} \\

\bottomrule
\end{tabular}

%% file: main.bbl
\begin{thebibliography}{80}
\expandafter\ifx\csname natexlab\endcsname\relax\def\natexlab#1{#1}\fi

\bibitem[{Abzianidze et~al.(2017)Abzianidze, Bjerva, Evang, Haagsma, van Noord, Ludmann, Nguyen, and Bos}]{abzianidze-etal-2017-parallel}
Lasha Abzianidze, Johannes Bjerva, Kilian Evang, Hessel Haagsma, Rik van Noord, Pierre Ludmann, Duc-Duy Nguyen, and Johan Bos. 2017.
\newblock \href {https://aclanthology.org/E17-2039} {The {P}arallel {M}eaning {B}ank: Towards a multilingual corpus of translations annotated with compositional meaning representations}.
\newblock In \emph{Proceedings of the 15th Conference of the {E}uropean Chapter of the Association for Computational Linguistics: Volume 2, Short Papers}, pages 242--247, Valencia, Spain. Association for Computational Linguistics.

\bibitem[{Agarwal et~al.(2021)Agarwal, Yurochkin, and Sun}]{agarwal2021sensitivity}
Mayank Agarwal, Mikhail Yurochkin, and Yuekai Sun. 2021.
\newblock \href {https://proceedings.neurips.cc/paper/2021/hash/ab73f542b6d60c4de151800b8abc0a6c-Abstract.html} {On sensitivity of meta-learning to support data}.
\newblock In \emph{Advances in Neural Information Processing Systems 34: Annual Conference on Neural Information Processing Systems 2021, NeurIPS 2021, December 6-14, 2021, virtual}, pages 20447--20460.

\bibitem[{Akaike(1998)}]{akaike1998information}
Hirotogu Akaike. 1998.
\newblock \href {https://link.springer.com/chapter/10.1007/978-1-4612-1694-0_15} {Information theory and an extension of the maximum likelihood principle}.
\newblock In \emph{Selected papers of Hirotugu Akaike}, pages 199--213. Springer.

\bibitem[{Alzahrani et~al.(2024)Alzahrani, Alyahya, Alnumay, AlRashed, Alsubaie, Almushayqih, Mirza, Alotaibi, Al-Twairesh, Alowisheq, Bari, and Khan}]{alzahrani2024benchmarks}
Norah Alzahrani, Hisham Alyahya, Yazeed Alnumay, Sultan AlRashed, Shaykhah Alsubaie, Yousef Almushayqih, Faisal Mirza, Nouf Alotaibi, Nora Al-Twairesh, Areeb Alowisheq, M~Saiful Bari, and Haidar Khan. 2024.
\newblock \href {https://aclanthology.org/2024.acl-long.744} {When benchmarks are targets: {R}evealing the sensitivity of large language model leaderboards}.
\newblock In \emph{Proceedings of the 62nd Annual Meeting of the Association for Computational Linguistics (Volume 1: Long Papers)}, pages 13787--13805, Bangkok, Thailand. Association for Computational Linguistics.

\bibitem[{Andonian et~al.(2023)Andonian, Anthony, Biderman, Black, Gali, Gao, Hallahan, Levy-Kramer, Leahy, Nestler, Parker, Pieler, Phang, Purohit, Schoelkopf, Stander, Songz, Tigges, Th\'{e}rien, Wang, and Weinbach}]{gpt-neox-library}
Alex Andonian, Quentin Anthony, Stella Biderman, Sid Black, Preetham Gali, Leo Gao, Eric Hallahan, Josh Levy-Kramer, Connor Leahy, Lucas Nestler, Kip Parker, Michael Pieler, Jason Phang, Shivanshu Purohit, Hailey Schoelkopf, Dashiell Stander, Tri Songz, Curt Tigges, Benjamin Th\'{e}rien, Phil Wang, and Samuel Weinbach. 2023.
\newblock \href {https://www.github.com/eleutherai/gpt-neox} {{GPT-NeoX}: {L}arge scale autoregressive language modeling in {PyTorch}}.

\bibitem[{Arnold et~al.(2024)Arnold, Holtorf, Sch\"{a}fer, and L\"{o}rch}]{arnold2024phase}
Julian Arnold, Flemming Holtorf, Frank Sch\"{a}fer, and Niels L\"{o}rch. 2024.
\newblock \href {https://arxiv.org/abs/2405.17088} {Phase transitions in the output distribution of large language models}.
\newblock \emph{ArXiv preprint 2405.17088}.

\bibitem[{Baum and Petrie(1966)}]{baum1966statistical}
Leonard~E. Baum and Ted Petrie. 1966.
\newblock \href {https://www.jstor.org/stable/2238772} {Statistical inference for probabilistic functions of finite state markov chains}.
\newblock \emph{The annals of mathematical statistics}, 37(6):1554--1563.

\bibitem[{Baum et~al.(1970)Baum, Petrie, Soules, and Weiss}]{baum1970maximization}
Leonard~E. Baum, Ted Petrie, George Soules, and Norman Weiss. 1970.
\newblock \href {https://www.jstor.org/stable/2239727} {A maximization technique occurring in the statistical analysis of probabilistic functions of markov chains}.
\newblock \emph{The annals of mathematical statistics}, 41(1):164--171.

\bibitem[{Belrose et~al.(2024)Belrose, Pope, Quirke, Mallen, and Fern}]{belrose2024neural}
Nora Belrose, Quintin Pope, Lucia Quirke, Alex Mallen, and Xiaoli Fern. 2024.
\newblock \href {https://arxiv.org/abs/2402.04362} {Neural networks learn statistics of increasing complexity}.
\newblock \emph{ArXiv preprint}, abs/2402.04362.

\bibitem[{Biderman et~al.(2022)Biderman, Bicheno, and Gao}]{biderman-etal-2022-datasheet}
Stella Biderman, Kieran Bicheno, and Leo Gao. 2022.
\newblock \href {https://arxiv.org/abs/2201.07311} {Datasheet for the {{Pile}}}.
\newblock \emph{ArXiv preprint 2201.07311}.

\bibitem[{Biderman et~al.(2023{\natexlab{a}})Biderman, Prashanth, Sutawika, Schoelkopf, Anthony, Purohit, and Raff}]{biderman2024emergent}
Stella Biderman, USVSN~Sai Prashanth, Lintang Sutawika, Hailey Schoelkopf, Quentin Anthony, Shivanshu Purohit, and Edward Raff. 2023{\natexlab{a}}.
\newblock \href {http://papers.nips.cc/paper\_files/paper/2023/hash/59404fb89d6194641c69ae99ecdf8f6d-Abstract-Conference.html} {Emergent and predictable memorization in large language models}.
\newblock In \emph{Advances in Neural Information Processing Systems 36: Annual Conference on Neural Information Processing Systems 2023, NeurIPS 2023, New Orleans, LA, USA, December 10 - 16, 2023}.

\bibitem[{Biderman et~al.(2023{\natexlab{b}})Biderman, Schoelkopf, Anthony, Bradley, O'Brien, Hallahan, Khan, Purohit, Prashanth, Raff, Skowron, Sutawika, and van~der Wal}]{biderman-etal-2023-pythia}
Stella Biderman, Hailey Schoelkopf, Quentin~Gregory Anthony, Herbie Bradley, Kyle O'Brien, Eric Hallahan, Mohammad~Aflah Khan, Shivanshu Purohit, USVSN~Sai Prashanth, Edward Raff, Aviya Skowron, Lintang Sutawika, and Oskar van~der Wal. 2023{\natexlab{b}}.
\newblock \href {https://proceedings.mlr.press/v202/biderman23a.html} {Pythia: {A} suite for analyzing large language models across training and scaling}.
\newblock In \emph{International Conference on Machine Learning, {ICML} 2023, 23-29 July 2023, Honolulu, Hawaii, {USA}}, volume 202 of \emph{Proceedings of Machine Learning Research}, pages 2397--2430. {PMLR}.

\bibitem[{Biderman et~al.(2024)Biderman, Schoelkopf, Sutawika, Gao, Tow, Abbasi, Aji, Ammanamanchi, Black, Clive, DiPofi, Etxaniz, Fattori, Forde, Foster, Hsu, Jaiswal, Lee, Li, Lovering, Muennighoff, Pavlick, Phang, Skowron, Tan, Tang, Wang, Winata, Yvon, and Zou}]{biderman-etal-2024-evalharness}
Stella Biderman, Hailey Schoelkopf, Lintang Sutawika, Leo Gao, Jonathan Tow, Baber Abbasi, Alham~Fikri Aji, Pawan~Sasanka Ammanamanchi, Sidney Black, Jordan Clive, Anthony DiPofi, Julen Etxaniz, Benjamin Fattori, Jessica~Zosa Forde, Charles Foster, Jeffrey Hsu, Mimansa Jaiswal, Wilson~Y. Lee, Haonan Li, Charles Lovering, Niklas Muennighoff, Ellie Pavlick, Jason Phang, Aviya Skowron, Samson Tan, Xiangru Tang, Kevin~A. Wang, Genta~Indra Winata, Fran\c{c}ois Yvon, and Andy Zou. 2024.
\newblock \href {https://arxiv.org/abs/2405.14782} {Lessons from the trenches on reproducible evaluation of language models}.
\newblock \emph{ArXiv preprint 2405.14782}.

\bibitem[{Bisk et~al.(2020)Bisk, Zellers, LeBras, Gao, and Choi}]{bisk-piqa}
Yonatan Bisk, Rowan Zellers, Ronan LeBras, Jianfeng Gao, and Yejin Choi. 2020.
\newblock \href {https://aaai.org/ojs/index.php/AAAI/article/view/6239} {{PIQA:} {R}easoning about physical commonsense in natural language}.
\newblock In \emph{The Thirty-Fourth {AAAI} Conference on Artificial Intelligence, {AAAI} 2020, The Thirty-Second Innovative Applications of Artificial Intelligence Conference, {IAAI} 2020, The Tenth {AAAI} Symposium on Educational Advances in Artificial Intelligence, {EAAI} 2020, New York, NY, USA, February 7-12, 2020}, pages 7432--7439. {AAAI} Press.

\bibitem[{Blodgett et~al.(2021)Blodgett, Lopez, Olteanu, Sim, and Wallach}]{blodgett-etal-2021-stereotyping}
Su~Lin Blodgett, Gilsinia Lopez, Alexandra Olteanu, Robert Sim, and Hanna Wallach. 2021.
\newblock \href {https://doi.org/10.18653/v1/2021.acl-long.81} {Stereotyping {N}orwegian salmon: An inventory of pitfalls in fairness benchmark datasets}.
\newblock In \emph{Proceedings of the 59th Annual Meeting of the Association for Computational Linguistics and the 11th International Joint Conference on Natural Language Processing (Volume 1: Long Papers)}, pages 1004--1015, Online. Association for Computational Linguistics.

\bibitem[{Brown et~al.(2020)Brown, Mann, Ryder, Subbiah, Kaplan, Dhariwal, Neelakantan, Shyam, Sastry, Askell, Agarwal, Herbert{-}Voss, Krueger, Henighan, Child, Ramesh, Ziegler, Wu, Winter, Hesse, Chen, Sigler, Litwin, Gray, Chess, Clark, Berner, McCandlish, Radford, Sutskever, and Amodei}]{brown-etal-2020-language}
Tom~B. Brown, Benjamin Mann, Nick Ryder, Melanie Subbiah, Jared Kaplan, Prafulla Dhariwal, Arvind Neelakantan, Pranav Shyam, Girish Sastry, Amanda Askell, Sandhini Agarwal, Ariel Herbert{-}Voss, Gretchen Krueger, Tom Henighan, Rewon Child, Aditya Ramesh, Daniel~M. Ziegler, Jeffrey Wu, Clemens Winter, Christopher Hesse, Mark Chen, Eric Sigler, Mateusz Litwin, Scott Gray, Benjamin Chess, Jack Clark, Christopher Berner, Sam McCandlish, Alec Radford, Ilya Sutskever, and Dario Amodei. 2020.
\newblock \href {https://proceedings.neurips.cc/paper/2020/hash/1457c0d6bfcb4967418bfb8ac142f64a-Abstract.html} {Language models are few-shot learners}.
\newblock In \emph{Advances in Neural Information Processing Systems 33: Annual Conference on Neural Information Processing Systems 2020, NeurIPS 2020, December 6-12, 2020, virtual}.

\bibitem[{Chang and Bergen(2022)}]{chang-bergen-2022-word}
Tyler~A. Chang and Benjamin~K. Bergen. 2022.
\newblock \href {https://doi.org/10.1162/tacl_a_00444} {Word acquisition in neural language models}.
\newblock \emph{Transactions of the Association for Computational Linguistics}, 10:1--16.

\bibitem[{Chowdhery et~al.(2023)Chowdhery, Narang, Devlin, Bosma, Mishra, Roberts, Barham, Chung, Sutton, Gehrmann et~al.}]{chowdhery2023palm}
Aakanksha Chowdhery, Sharan Narang, Jacob Devlin, Maarten Bosma, Gaurav Mishra, Adam Roberts, Paul Barham, Hyung~Won Chung, Charles Sutton, Sebastian Gehrmann, et~al. 2023.
\newblock \href {https://jmlr.org/papers/v24/22-1144.html} {{PaLM}: {S}caling language modeling with pathways}.
\newblock \emph{Journal of Machine Learning Research}, 24(240):1--113.

\bibitem[{Chung et~al.(2024)Chung, Hong, An, Thorne, and Yun}]{chung2024stable}
Woojin Chung, Jiwoo Hong, Na~Min An, James Thorne, and Se-Young Yun. 2024.
\newblock \href {https://aclanthology.org/2024.emnlp-main.606/} {Stable language model pre-training by reducing embedding variability}.
\newblock In \emph{Proceedings of the 2024 Conference on Empirical Methods in Natural Language Processing}, pages 10852--10863, Miami, Florida, USA. Association for Computational Linguistics.

\bibitem[{Clark et~al.(2018)Clark, Cowhey, Etzioni, Khot, Sabharwal, Schoenick, and Tafjord}]{clark2018arc}
Peter Clark, Isaac Cowhey, Oren Etzioni, Tushar Khot, Ashish Sabharwal, Carissa Schoenick, and Oyvind Tafjord. 2018.
\newblock \href {https://arxiv.org/abs/1803.05457} {Think you have solved question answering? {T}ry {ARC}, the {AI2} reasoning challenge}.
\newblock \emph{ArXiv preprint 1803.05457}.

\bibitem[{Cohen(1960)}]{Cohen1960}
Jacob Cohen. 1960.
\newblock \href {https://journals.sagepub.com/doi/pdf/10.1177/001316446002000104} {A coefficient of agreement for nominal scales}.
\newblock \emph{Educational and Psychological Measurement}, 20:37--46.

\bibitem[{D'Amour et~al.(2022)D'Amour, Heller, Moldovan, Adlam, Alipanahi, Beutel, Chen, Deaton, Eisenstein, Hoffman et~al.}]{damour2022underspecification}
Alexander D'Amour, Katherine Heller, Dan Moldovan, Ben Adlam, Babak Alipanahi, Alex Beutel, Christina Chen, Jonathan Deaton, Jacob Eisenstein, Matthew~D Hoffman, et~al. 2022.
\newblock \href {https://www.jmlr.org/papers/v23/20-1335.html} {Underspecification presents challenges for credibility in modern machine learning}.
\newblock \emph{Journal of Machine Learning Research}, 23(226):1--61.

\bibitem[{Delobelle et~al.(2024)Delobelle, Attanasio, Nozza, Blodgett, and Talat}]{delobelle-etal-2024-metrics}
Pieter Delobelle, Giuseppe Attanasio, Debora Nozza, Su~Lin Blodgett, and Zeerak Talat. 2024.
\newblock \href {https://doi.org/10.18653/v1/2024.emnlp-main.1207} {Metrics for what, metrics for whom: {A}ssessing actionability of bias evaluation metrics in {NLP}}.
\newblock In \emph{Proceedings of the 2024 Conference on Empirical Methods in Natural Language Processing}, pages 21669--21691, Miami, Florida, USA. Association for Computational Linguistics.

\bibitem[{Devlin et~al.(2019)Devlin, Chang, Lee, and Toutanova}]{devlin-etal-2019-bert}
Jacob Devlin, Ming-Wei Chang, Kenton Lee, and Kristina Toutanova. 2019.
\newblock \href {https://doi.org/10.18653/v1/N19-1423} {{BERT}: Pre-training of deep bidirectional transformers for language understanding}.
\newblock In \emph{Proceedings of the 2019 Conference of the North {A}merican Chapter of the Association for Computational Linguistics: Human Language Technologies, Volume 1 (Long and Short Papers)}, pages 4171--4186, Minneapolis, Minnesota. Association for Computational Linguistics.

\bibitem[{Diehl~Martinez et~al.(2024)Diehl~Martinez, Lesci, and Buttery}]{diehl-martinez-etal-2024-tending}
Richard Diehl~Martinez, Pietro Lesci, and Paula Buttery. 2024.
\newblock \href {https://aclanthology.org/2024.findings-emnlp.187/} {Tending towards stability: {C}onvergence challenges in small language models}.
\newblock In \emph{Findings of the Association for Computational Linguistics: EMNLP 2024}, pages 3275--3286, Miami, Florida, USA. Association for Computational Linguistics.

\bibitem[{Dodge et~al.(2020)Dodge, Ilharco, Schwartz, Farhadi, Hajishirzi, and Smith}]{dodge-etal-2020-finetuning}
Jesse Dodge, Gabriel Ilharco, Roy Schwartz, Ali Farhadi, Hannaneh Hajishirzi, and Noah Smith. 2020.
\newblock \href {https://arxiv.org/abs/2002.06305} {Fine-tuning pretrained language models: {W}eight initializations, data orders, and early stopping}.
\newblock \emph{ArXiv preprint 2002.06305}.

\bibitem[{Du and Nguyen(2023)}]{du-nguyen-2023-measuring}
Yupei Du and Dong Nguyen. 2023.
\newblock \href {https://doi.org/10.18653/v1/2023.acl-long.342} {Measuring the instability of fine-tuning}.
\newblock In \emph{Proceedings of the 61st Annual Meeting of the Association for Computational Linguistics (Volume 1: Long Papers)}, pages 6209--6230, Toronto, Canada. Association for Computational Linguistics.

\bibitem[{Fisher(1970)}]{fisher1970statistical}
Ronald~A. Fisher. 1970.
\newblock \href {https://link.springer.com/chapter/10.1007/978-1-4612-4380-9_6} {\emph{Statistical methods for research workers.}}, 14. ed. edition.
\newblock Oliver and Boyd, Edinburgh.

\bibitem[{Gao et~al.(2021)Gao, Biderman, Black, Golding, Hoppe, Foster, Phang, He, Thite, Nabeshima, Presser, and Leahy}]{gao-etal-2020-pile}
Leo Gao, Stella Biderman, Sid Black, Laurence Golding, Travis Hoppe, Charles Foster, Jason Phang, Horace He, Anish Thite, Noa Nabeshima, Shawn Presser, and Connor Leahy. 2021.
\newblock \href {https://arxiv.org/abs/2101.00027} {The {{Pile}}: {{An 800GB}} dataset of diverse text for language modeling}.
\newblock \emph{ArXiv preprint 2101.00027}.

\bibitem[{Gao et~al.(2024)Gao, Tow, Abbasi, Biderman, Black, DiPofi, Foster, Golding, Hsu, Le~Noac'h, Li, McDonell, Muennighoff, Ociepa, Phang, Reynolds, Schoelkopf, Skowron, Sutawika, Tang, Thite, Wang, Wang, and Zou}]{eval-harness}
Leo Gao, Jonathan Tow, Baber Abbasi, Stella Biderman, Sid Black, Anthony DiPofi, Charles Foster, Laurence Golding, Jeffrey Hsu, Alain Le~Noac'h, Haonan Li, Kyle McDonell, Niklas Muennighoff, Chris Ociepa, Jason Phang, Laria Reynolds, Hailey Schoelkopf, Aviya Skowron, Lintang Sutawika, Eric Tang, Anish Thite, Ben Wang, Kevin Wang, and Andy Zou. 2024.
\newblock \href {https://zenodo.org/records/12608602} {A framework for few-shot language model evaluation}.

\bibitem[{Godey et~al.(2024)Godey, \'{E}ric de~la Clergerie, and Sagot}]{godey2024small}
Nathan Godey, \'{E}ric de~la Clergerie, and Beno\^{\i}t Sagot. 2024.
\newblock \href {https://arxiv.org/abs/2404.07647} {Why do small language models underperform? {S}tudying language model saturation via the softmax bottleneck}.
\newblock \emph{ArXiv preprint 2404.07647}.

\bibitem[{Goldfarb-Tarrant et~al.(2024)Goldfarb-Tarrant, Rodriguez, Dwivedi-Yu, and Lewis}]{goldfarbtarrant2024multicontrievers}
Seraphina Goldfarb-Tarrant, Pedro Rodriguez, Jane Dwivedi-Yu, and Patrick Lewis. 2024.
\newblock \href {https://doi.org/10.18653/v1/2024.blackboxnlp-1.8} {{M}ulti{C}ontrievers: {A}nalysis of dense retrieval representations}.
\newblock In \emph{Proceedings of the 7th BlackboxNLP Workshop: Analyzing and Interpreting Neural Networks for NLP}, pages 118--139, Miami, Florida, US. Association for Computational Linguistics.

\bibitem[{Gould et~al.(2024)Gould, Ong, Ogden, and Conmy}]{gould2024successor}
Rhys Gould, Euan Ong, George Ogden, and Arthur Conmy. 2024.
\newblock \href {https://openreview.net/forum?id=kvcbV8KQsi} {Successor heads: {R}ecurring, interpretable attention heads in the wild}.
\newblock In \emph{The Twelfth International Conference on Learning Representations}.

\bibitem[{Gundersen et~al.(2022)Gundersen, Coakley, Kirkpatrick, and Gil}]{gundersen2023sources}
Odd~Erik Gundersen, Kevin Coakley, Christine Kirkpatrick, and Yolanda Gil. 2022.
\newblock \href {https://arxiv.org/abs/2204.07610} {Sources of irreproducibility in machine learning: {A} review}.
\newblock \emph{ArXiv preprint 2204.07610}.

\bibitem[{Hoffmann et~al.(2022)Hoffmann, Borgeaud, Mensch, Buchatskaya, Cai, Rutherford, de~Las~Casas, Hendricks, Welbl, Clark, Hennigan, Noland, Millican, van~den Driessche, Damoc, Guy, Osindero, Simonyan, Elsen, Vinyals, Rae, and Sifre}]{hoffmann-etal-2022-training}
Jordan Hoffmann, Sebastian Borgeaud, Arthur Mensch, Elena Buchatskaya, Trevor Cai, Eliza Rutherford, Diego de~Las~Casas, Lisa~Anne Hendricks, Johannes Welbl, Aidan Clark, Tom Hennigan, Eric Noland, Katherine Millican, George van~den Driessche, Bogdan Damoc, Aurelia Guy, Simon Osindero, Karen Simonyan, Erich Elsen, Oriol Vinyals, Jack~W. Rae, and Laurent Sifre. 2022.
\newblock \href {http://papers.nips.cc/paper\_files/paper/2022/hash/c1e2faff6f588870935f114ebe04a3e5-Abstract-Conference.html} {An empirical analysis of compute-optimal large language model training}.
\newblock In \emph{Advances in Neural Information Processing Systems 35: Annual Conference on Neural Information Processing Systems 2022, NeurIPS 2022, New Orleans, LA, USA, November 28 - December 9, 2022}.

\bibitem[{Hu et~al.(2023)Hu, Chen, Saphra, and Cho}]{hu2023latent}
Michael~Y. Hu, Angelica Chen, Naomi Saphra, and Kyunghyun Cho. 2023.
\newblock \href {https://openreview.net/forum?id=NE2xXWo0LF} {Latent state models of training dynamics}.
\newblock \emph{Transactions on Machine Learning Research}.

\bibitem[{Hu et~al.(2024)Hu, Kyrychenko, Rathje, Collier, van~der Linden, and Roozenbeek}]{hu2024generative}
Tiancheng Hu, Yara Kyrychenko, Steve Rathje, Nigel Collier, Sander van~der Linden, and Jon Roozenbeek. 2024.
\newblock \href {https://doi.org/10.1038/s43588-024-00741-1} {Generative language models exhibit social identity biases}.
\newblock \emph{Nature Computational Science}, 5(1):65--75.

\bibitem[{Jumelet et~al.(2024)Jumelet, Bylinina, Zuidema, and Szymanik}]{jumelet2024black}
Jaap Jumelet, Lisa Bylinina, Willem Zuidema, and Jakub Szymanik. 2024.
\newblock \href {https://arxiv.org/abs/2407.02136} {Black big boxes: {D}o language models hide a theory of adjective order?}
\newblock \emph{ArXiv preprint 2407.02136}.

\bibitem[{Kaplan et~al.(2020)Kaplan, McCandlish, Henighan, Brown, Chess, Child, Gray, Radford, Wu, and Amodei}]{kaplan2020scaling}
Jared Kaplan, Sam McCandlish, Tom Henighan, Tom~B Brown, Benjamin Chess, Rewon Child, Scott Gray, Alec Radford, Jeffrey Wu, and Dario Amodei. 2020.
\newblock \href {https://arxiv.org/abs/2001.08361} {Scaling laws for neural language models}.
\newblock \emph{ArXiv preprint 2001.08361}.

\bibitem[{Karamcheti et~al.(2021)Karamcheti, Orr, Bolton, Zhang, Goel, Narayan, Bommasani, Narayanan, Hashimoto, Jurafsky, Manning, Potts, R\'{e}, and Liang}]{karamcheti2021mistral}
Siddharth Karamcheti, Laurel Orr, Jason Bolton, Tianyi Zhang, Karan Goel, Avanika Narayan, Rishi Bommasani, Deepak Narayanan, Tatsunori Hashimoto, Dan Jurafsky, Christopher~D. Manning, Christopher Potts, Christopher R\'{e}, and Percy Liang. 2021.
\newblock \href {https://crfm.stanford.edu/2021/08/26/mistral.html} {Mistral--{A} journey towards reproducible language model training}.

\bibitem[{Knyazev and Argentati(2002)}]{knyazev2002ssa}
Andrew~V. Knyazev and Merico~E. Argentati. 2002.
\newblock \href {https://epubs.siam.org/doi/10.1137/S1064827500377332} {Principal angles between subspaces in an {A}-based scalar product: {A}lgorithms and perturbation estimates}.
\newblock \emph{SIAM Journal on Scientific Computing}, 23(6):2008--2040.

\bibitem[{Lang(1995)}]{lang95news}
Ken Lang. 1995.
\newblock \href {https://dl.acm.org/doi/10.5555/3091622.3091662} {Newsweeder: {L}earning to filter netnews}.
\newblock In \emph{Proceedings of the Twelfth International Conference on International Conference on Machine Learning}, ICML'95, page 331–339, San Francisco, CA, USA. Morgan Kaufmann Publishers Inc.

\bibitem[{Lesci et~al.(2024)Lesci, Meister, Hofmann, Vlachos, and Pimentel}]{lesci-etal-2024-causal}
Pietro Lesci, Clara Meister, Thomas Hofmann, Andreas Vlachos, and Tiago Pimentel. 2024.
\newblock \href {https://aclanthology.org/2024.acl-long.834} {Causal estimation of memorisation profiles}.
\newblock In \emph{Proceedings of the 62nd Annual Meeting of the Association for Computational Linguistics (Volume 1: Long Papers)}, pages 15616--15635, Bangkok, Thailand. Association for Computational Linguistics.

\bibitem[{Levesque et~al.(2012)Levesque, Davis, and Morgenstern}]{wsc}
Hector~J. Levesque, Ernest Davis, and Leora Morgenstern. 2012.
\newblock \href {https://dl.acm.org/doi/10.5555/3031843.3031909} {The {Winograd} schema challenge}.
\newblock In \emph{Proceedings of the Thirteenth International Conference on Principles of Knowledge Representation and Reasoning}, KR'12, page 552–561. AAAI Press.

\bibitem[{Liu et~al.(2020)Liu, Cui, Liu, Huang, Wang, and Zhang}]{logiqa}
Jian Liu, Leyang Cui, Hanmeng Liu, Dandan Huang, Yile Wang, and Yue Zhang. 2020.
\newblock \href {https://doi.org/10.24963/ijcai.2020/501} {Logi{QA}: {A} challenge dataset for machine reading comprehension with logical reasoning}.
\newblock In \emph{Proceedings of the Twenty-Ninth International Joint Conference on Artificial Intelligence, {IJCAI} 2020}, pages 3622--3628. ijcai.org.

\bibitem[{Madaan et~al.(2024)Madaan, Singh, Schaeffer, Poulton, Koyejo, Stenetorp, Narang, and Hupkes}]{madaan2024quantifying}
Lovish Madaan, Aaditya~K. Singh, Rylan Schaeffer, Andrew Poulton, Sanmi Koyejo, Pontus Stenetorp, Sharan Narang, and Dieuwke Hupkes. 2024.
\newblock \href {https://arxiv.org/abs/2406.10229} {Quantifying variance in evaluation benchmarks}.
\newblock \emph{ArXiv preprint 2406.10229}.

\bibitem[{McCoy et~al.(2020)McCoy, Min, and Linzen}]{mccoy-etal-2020-berts}
R.~Thomas McCoy, Junghyun Min, and Tal Linzen. 2020.
\newblock \href {https://doi.org/10.18653/v1/2020.blackboxnlp-1.21} {{BERT}s of a feather do not generalize together: Large variability in generalization across models with similar test set performance}.
\newblock In \emph{Proceedings of the Third BlackboxNLP Workshop on Analyzing and Interpreting Neural Networks for NLP}, pages 217--227, Online. Association for Computational Linguistics.

\bibitem[{McDougall et~al.(2023)McDougall, Conmy, Rushing, McGrath, and Nanda}]{mcdougall2023copy}
Callum McDougall, Arthur Conmy, Cody Rushing, Thomas McGrath, and Neel Nanda. 2023.
\newblock \href {https://arxiv.org/abs/2310.04625} {Copy suppression: {C}omprehensively understanding an attention head}.
\newblock \emph{ArXiv preprint 2310.04625}.

\bibitem[{Meister et~al.(2023)Meister, Stokowiec, Pimentel, Yu, Rimell, and Kuncoro}]{meister-etal-2023-natural}
Clara Meister, Wojciech Stokowiec, Tiago Pimentel, Lei Yu, Laura Rimell, and Adhiguna Kuncoro. 2023.
\newblock \href {https://doi.org/10.18653/v1/2023.acl-short.22} {A natural bias for language generation models}.
\newblock In \emph{Proceedings of the 61st Annual Meeting of the Association for Computational Linguistics (Volume 2: Short Papers)}, pages 243--255, Toronto, Canada. Association for Computational Linguistics.

\bibitem[{Michaelov and Bergen(2023)}]{michaelov-bergen-2023-emergent}
James Michaelov and Ben Bergen. 2023.
\newblock \href {https://doi.org/10.18653/v1/2023.findings-emnlp.973} {Emergent inabilities? {I}nverse scaling over the course of pretraining}.
\newblock In \emph{Findings of the Association for Computational Linguistics: EMNLP 2023}, pages 14607--14615, Singapore. Association for Computational Linguistics.

\bibitem[{Mosbach et~al.(2021)Mosbach, Andriushchenko, and Klakow}]{mosbach2021on}
Marius Mosbach, Maksym Andriushchenko, and Dietrich Klakow. 2021.
\newblock \href {https://openreview.net/forum?id=nzpLWnVAyah} {On the stability of fine-tuning {BERT:} {M}isconceptions, explanations, and strong baselines}.
\newblock In \emph{9th International Conference on Learning Representations, {ICLR} 2021, Virtual Event, Austria, May 3-7, 2021}.

\bibitem[{M{\"u}ller-Eberstein et~al.(2023)M{\"u}ller-Eberstein, van~der Goot, Plank, and Titov}]{muller-eberstein-etal-2023-subspace}
Max M{\"u}ller-Eberstein, Rob van~der Goot, Barbara Plank, and Ivan Titov. 2023.
\newblock \href {https://doi.org/10.18653/v1/2023.findings-emnlp.879} {Subspace chronicles: {H}ow linguistic information emerges, shifts and interacts during language model training}.
\newblock In \emph{Findings of the Association for Computational Linguistics: EMNLP 2023}, pages 13190--13208, Singapore. Association for Computational Linguistics.

\bibitem[{Nangia et~al.(2020)Nangia, Vania, Bhalerao, and Bowman}]{nangia-etal-2020-crows}
Nikita Nangia, Clara Vania, Rasika Bhalerao, and Samuel~R. Bowman. 2020.
\newblock \href {https://doi.org/10.18653/v1/2020.emnlp-main.154} {{C}row{S}-pairs: A challenge dataset for measuring social biases in masked language models}.
\newblock In \emph{Proceedings of the 2020 Conference on Empirical Methods in Natural Language Processing (EMNLP)}, pages 1953--1967, Online. Association for Computational Linguistics.

\bibitem[{N{\'e}v{\'e}ol et~al.(2022)N{\'e}v{\'e}ol, Dupont, Bezan{\c{c}}on, and Fort}]{neveol-etal-2022-french}
Aur{\'e}lie N{\'e}v{\'e}ol, Yoann Dupont, Julien Bezan{\c{c}}on, and Kar{\"e}n Fort. 2022.
\newblock \href {https://doi.org/10.18653/v1/2022.acl-long.583} {{F}rench {C}row{S}-pairs: Extending a challenge dataset for measuring social bias in masked language models to a language other than {E}nglish}.
\newblock In \emph{Proceedings of the 60th Annual Meeting of the Association for Computational Linguistics (Volume 1: Long Papers)}, pages 8521--8531, Dublin, Ireland. Association for Computational Linguistics.

\bibitem[{Olsson et~al.(2022)Olsson, Elhage, Nanda, Joseph, DasSarma, Henighan, Mann, Askell, Bai, Chen et~al.}]{olsson2022context}
Catherine Olsson, Nelson Elhage, Neel Nanda, Nicholas Joseph, Nova DasSarma, Tom Henighan, Ben Mann, Amanda Askell, Yuntao Bai, Anna Chen, et~al. 2022.
\newblock \href {https://arxiv.org/abs/2209.11895} {In-context learning and induction heads}.
\newblock \emph{ArXiv preprint 2209.11895}.

\bibitem[{Paperno et~al.(2016)Paperno, Kruszewski, Lazaridou, Pham, Bernardi, Pezzelle, Baroni, Boleda, and Fern{\'a}ndez}]{paperno-etal-2016-lambada}
Denis Paperno, Germ{\'a}n Kruszewski, Angeliki Lazaridou, Ngoc~Quan Pham, Raffaella Bernardi, Sandro Pezzelle, Marco Baroni, Gemma Boleda, and Raquel Fern{\'a}ndez. 2016.
\newblock \href {https://doi.org/10.18653/v1/P16-1144} {The {LAMBADA} dataset: {W}ord prediction requiring a broad discourse context}.
\newblock In \emph{Proceedings of the 54th Annual Meeting of the Association for Computational Linguistics (Volume 1: Long Papers)}, pages 1525--1534, Berlin, Germany. Association for Computational Linguistics.

\bibitem[{Pecher et~al.(2024)Pecher, Srba, and Bielikova}]{pecher-etal-2024-survey}
Branislav Pecher, Ivan Srba, and Maria Bielikova. 2024.
\newblock \href {https://doi.org/10.1145/3691339} {A survey on stability of learning with limited labelled data and its sensitivity to the effects of randomness}.
\newblock \emph{ACM Comput. Surv.}

\bibitem[{Pham et~al.(2020)Pham, Qian, Wang, Lutellier, Rosenthal, Tan, Yu, and Nagappan}]{pham2020problems}
Hung~Viet Pham, Shangshu Qian, Jiannan Wang, Thibaud Lutellier, Jonathan Rosenthal, Lin Tan, Yaoliang Yu, and Nachiappan Nagappan. 2020.
\newblock \href {https://dl.acm.org/doi/10.1145/3324884.3416545} {Problems and opportunities in training deep learning software systems: {A}n analysis of variance}.
\newblock In \emph{Proceedings of the 35th IEEE/ACM international conference on automated software engineering}, pages 771--783.

\bibitem[{Pimentel et~al.(2020)Pimentel, Valvoda, Maudslay, Zmigrod, Williams, and Cotterell}]{pimentel-etal-2020-information}
Tiago Pimentel, Josef Valvoda, Rowan~Hall Maudslay, Ran Zmigrod, Adina Williams, and Ryan Cotterell. 2020.
\newblock \href {https://doi.org/10.18653/v1/2020.acl-main.420} {Information-theoretic probing for linguistic structure}.
\newblock In \emph{Proceedings of the 58th Annual Meeting of the Association for Computational Linguistics}, pages 4609--4622, Online. Association for Computational Linguistics.

\bibitem[{Pradhan et~al.(2013)Pradhan, Moschitti, Xue, Ng, Bj{\"o}rkelund, Uryupina, Zhang, and Zhong}]{pradhan-etal-2013-towards}
Sameer Pradhan, Alessandro Moschitti, Nianwen Xue, Hwee~Tou Ng, Anders Bj{\"o}rkelund, Olga Uryupina, Yuchen Zhang, and Zhi Zhong. 2013.
\newblock \href {https://aclanthology.org/W13-3516} {Towards robust linguistic analysis using {O}nto{N}otes}.
\newblock In \emph{Proceedings of the Seventeenth Conference on Computational Natural Language Learning}, pages 143--152, Sofia, Bulgaria. Association for Computational Linguistics.

\bibitem[{Radford et~al.(2019)Radford, Wu, Child, Luan, Amodei, and Sutskever}]{radford2019language}
Alec Radford, Jeff Wu, Rewon Child, David Luan, Dario Amodei, and Ilya Sutskever. 2019.
\newblock \href {https://cdn.openai.com/better-language-models/language\%5Fmodels\%5Fare\%5Funsupervised\%5Fmultitask\%5Flearners.pdf} {Language models are unsupervised multitask learners}.

\bibitem[{Sakaguchi et~al.(2020)Sakaguchi, Bras, Bhagavatula, and Choi}]{sakaguchi-etal-wino}
Keisuke Sakaguchi, Ronan~Le Bras, Chandra Bhagavatula, and Yejin Choi. 2020.
\newblock \href {https://aaai.org/ojs/index.php/AAAI/article/view/6399} {{WinoGrande}: {A}n adversarial winograd schema challenge at scale}.
\newblock In \emph{The Thirty-Fourth {AAAI} Conference on Artificial Intelligence, {AAAI} 2020, The Thirty-Second Innovative Applications of Artificial Intelligence Conference, {IAAI} 2020, The Tenth {AAAI} Symposium on Educational Advances in Artificial Intelligence, {EAAI} 2020, New York, NY, USA, February 7-12, 2020}, pages 8732--8740. {AAAI} Press.

\bibitem[{Saphra and Lopez(2019)}]{saphra-lopez-2019-understanding}
Naomi Saphra and Adam Lopez. 2019.
\newblock \href {https://doi.org/10.18653/v1/N19-1329} {Understanding learning dynamics of language models with {SVCCA}}.
\newblock In \emph{Proceedings of the 2019 Conference of the North {A}merican Chapter of the Association for Computational Linguistics: Human Language Technologies, Volume 1 (Long and Short Papers)}, pages 3257--3267, Minneapolis, Minnesota. Association for Computational Linguistics.

\bibitem[{Schwarz(1978)}]{schwarz1978estimating}
Gideon Schwarz. 1978.
\newblock \href {https://www.jstor.org/stable/2958889} {Estimating the dimension of a model}.
\newblock \emph{The annals of statistics}, pages 461--464.

\bibitem[{Sellam et~al.(2022)Sellam, Yadlowsky, Tenney, Wei, Saphra, D'Amour, Linzen, Bastings, Turc, Eisenstein, Das, and Pavlick}]{sellam2022the}
Thibault Sellam, Steve Yadlowsky, Ian Tenney, Jason Wei, Naomi Saphra, Alexander D'Amour, Tal Linzen, Jasmijn Bastings, Iulia~Raluca Turc, Jacob Eisenstein, Dipanjan Das, and Ellie Pavlick. 2022.
\newblock \href {https://openreview.net/forum?id=K0E\_F0gFDgA} {The {MultiBERTs}: {BERT} reproductions for robustness analysis}.
\newblock In \emph{The Tenth International Conference on Learning Representations, {ICLR} 2022, Virtual Event, April 25-29, 2022}.

\bibitem[{Silveira et~al.(2014)Silveira, Dozat, de~Marneffe, Bowman, Connor, Bauer, and Manning}]{silveira-etal-2014-gold}
Natalia Silveira, Timothy Dozat, Marie-Catherine de~Marneffe, Samuel Bowman, Miriam Connor, John Bauer, and Chris Manning. 2014.
\newblock \href {http://www.lrec-conf.org/proceedings/lrec2014/pdf/1089_Paper.pdf} {A gold standard dependency corpus for {E}nglish}.
\newblock In \emph{Proceedings of the Ninth International Conference on Language Resources and Evaluation ({LREC}'14)}, pages 2897--2904, Reykjavik, Iceland. European Language Resources Association (ELRA).

\bibitem[{Smith et~al.(2022)Smith, Patwary, Norick, LeGresley, Rajbhandari, Casper, Liu, Prabhumoye, Zerveas, Korthikanti, Zhang, Child, Aminabadi, Bernauer, Song, Shoeybi, He, Houston, Tiwary, and Catanzaro}]{smith2022using}
Shaden Smith, Mostofa Patwary, Brandon Norick, Patrick LeGresley, Samyam Rajbhandari, Jared Casper, Zhun Liu, Shrimai Prabhumoye, George Zerveas, Vijay Korthikanti, Elton Zhang, Rewon Child, Reza~Yazdani Aminabadi, Julie Bernauer, Xia Song, Mohammad Shoeybi, Yuxiong He, Michael Houston, Saurabh Tiwary, and Bryan Catanzaro. 2022.
\newblock \href {https://arxiv.org/abs/2201.11990} {Using {DeepSpeed} and {Megatron} to train {Megatron-Turing} {NLG} {530B}, a large-scale generative language model}.
\newblock \emph{Arxiv preprint 2201.11990}.

\bibitem[{Socher et~al.(2013)Socher, Perelygin, Wu, Chuang, Manning, Ng, and Potts}]{socher-etal-2013-recursive}
Richard Socher, Alex Perelygin, Jean Wu, Jason Chuang, Christopher~D. Manning, Andrew Ng, and Christopher Potts. 2013.
\newblock \href {https://aclanthology.org/D13-1170} {Recursive deep models for semantic compositionality over a sentiment treebank}.
\newblock In \emph{Proceedings of the 2013 Conference on Empirical Methods in Natural Language Processing}, pages 1631--1642, Seattle, Washington, USA. Association for Computational Linguistics.

\bibitem[{Srivastava et~al.(2023)Srivastava, Rastogi, Rao, Shoeb, Abid, Fisch, Brown, Santoro, Gupta, Garriga-Alonso, Kluska, Lewkowycz, Agarwal, Power, Ray, Warstadt, Kocurek, Safaya, Tazarv, Xiang, Parrish, Nie, Hussain, Askell, Dsouza, Slone, Rahane, Iyer, Andreassen, Madotto, Santilli, Stuhlm{\"u}ller, Dai, La, Lampinen, Zou, Jiang, Chen, Vuong, Gupta, Gottardi, Norelli, Venkatesh, Gholamidavoodi, Tabassum, Menezes, Kirubarajan, Mullokandov, Sabharwal, Herrick, Efrat, Erdem, Karaka{\c{s}}, Roberts, Loe, Zoph, Bojanowski, {\"O}zyurt, Hedayatnia, Neyshabur, Inden, Stein, Ekmekci, Lin, Howald, Orinion, Diao, Dour, Stinson, Argueta, Ferri, Singh, Rathkopf, Meng, Baral, Wu, Callison-Burch, Waites, Voigt, Manning, Potts, Ramirez, Rivera, Siro, Raffel, Ashcraft, Garbacea, Sileo, Garrette, Hendrycks, Kilman, Roth, Freeman, Khashabi, Levy, Gonz{\'a}lez, Perszyk, Hernandez, Chen, Ippolito, Gilboa, Dohan, Drakard, Jurgens, Datta, Ganguli, Emelin, Kleyko, Yuret, Chen, Tam, Hupkes, Misra, Buzan, Mollo, Yang, Lee,
  Schrader, Shutova, Cubuk, Segal, Hagerman, Barnes, Donoway, Pavlick, Rodol{\`a}, Lam, Chu, Tang, Erdem, Chang, Chi, Dyer, Jerzak, Kim, Manyasi, Zheltonozhskii, Xia, Siar, Mart{\'\i}nez-Plumed, Happ{\'e}, Chollet, Rong, Mishra, Winata, de~Melo, Kruszewski, Parascandolo, Mariani, Wang, Jaimovitch-Lopez, Betz, Gur-Ari, Galijasevic, Kim, Rashkin, Hajishirzi, Mehta, Bogar, Shevlin, Schuetze, Yakura, Zhang, Wong, Ng, Noble, Jumelet, Geissinger, Kernion, Hilton, Lee, Fisac, Simon, Koppel, Zheng, Zou, Kocon, Thompson, Wingfield, Kaplan, Radom, Sohl-Dickstein, Phang, Wei, Yosinski, Novikova, Bosscher, Marsh, Kim, Taal, Engel, Alabi, Xu, Song, Tang, Waweru, Burden, Miller, Balis, Batchelder, Berant, Frohberg, Rozen, Hernandez-Orallo, Boudeman, Guerr, Jones, Tenenbaum, Rule, Chua, Kanclerz, Livescu, Krauth, Gopalakrishnan, Ignatyeva, Markert, Dhole, Gimpel, Omondi, Mathewson, Chiafullo, Shkaruta, Shridhar, McDonell, Richardson, Reynolds, Gao, Zhang, Dugan, Qin, Contreras-Ochando, Morency, Moschella, Lam, Noble,
  Schmidt, He, Oliveros-Col{\'o}n, Metz, Senel, Bosma, Sap, Hoeve, Farooqi, Faruqui, Mazeika, Baturan, Marelli, Maru, Ramirez-Quintana, Tolkiehn, Giulianelli, Lewis, Potthast, Leavitt, Hagen, Schubert, Baitemirova, Arnaud, McElrath, Yee, Cohen, Gu, Ivanitskiy, Starritt, Strube, Sw{\k{e}}drowski, Bevilacqua, Yasunaga, Kale, Cain, Xu, Suzgun, Walker, Tiwari, Bansal, Aminnaseri, Geva, Gheini, T, Peng, Chi, Lee, Krakover, Cameron, Roberts, Doiron, Martinez, Nangia, Deckers, Muennighoff, Keskar, Iyer, Constant, Fiedel, Wen, Zhang, Agha, Elbaghdadi, Levy, Evans, Casares, Doshi, Fung, Liang, Vicol, Alipoormolabashi, Liao, Liang, Chang, Eckersley, Htut, Hwang, Mi{\l}kowski, Patil, Pezeshkpour, Oli, Mei, Lyu, Chen, Banjade, Rudolph, Gabriel, Habacker, Risco, Milli{\`e}re, Garg, Barnes, Saurous, Arakawa, Raymaekers, Frank, Sikand, Novak, Sitelew, Bras, Liu, Jacobs, Zhang, Salakhutdinov, Chi, Lee, Stovall, Teehan, Yang, Singh, Mohammad, Anand, Dillavou, Shleifer, Wiseman, Gruetter, Bowman, Schoenholz, Han, Kwatra, Rous,
  Ghazarian, Ghosh, Casey, Bischoff, Gehrmann, Schuster, Sadeghi, Hamdan, Zhou, Srivastava, Shi, Singh, Asaadi, Gu, Pachchigar, Toshniwal, Upadhyay, Debnath, Shakeri, Thormeyer, Melzi, Reddy, Makini, Lee, Torene, Hatwar, Dehaene, Divic, Ermon, Biderman, Lin, Prasad, Piantadosi, Shieber, Misherghi, Kiritchenko, Mishra, Linzen, Schuster, Li, Yu, Ali, Hashimoto, Wu, Desbordes, Rothschild, Phan, Wang, Nkinyili, Schick, Kornev, Tunduny, Gerstenberg, Chang, Neeraj, Khot, Shultz, Shaham, Misra, Demberg, Nyamai, Raunak, Ramasesh, vinay~uday prabhu, Padmakumar, Srikumar, Fedus, Saunders, Zhang, Vossen, Ren, Tong, Zhao, Wu, Shen, Yaghoobzadeh, Lakretz, Song, Bahri, Choi, Yang, Hao, Chen, Belinkov, Hou, Hou, Bai, Seid, Zhao, Wang, Wang, Wang, and Wu}]{srivastava2023beyond}
Aarohi Srivastava, Abhinav Rastogi, Abhishek Rao, Abu Awal~Md Shoeb, Abubakar Abid, Adam Fisch, Adam~R. Brown, Adam Santoro, Aditya Gupta, Adri{\`a} Garriga-Alonso, Agnieszka Kluska, Aitor Lewkowycz, Akshat Agarwal, Alethea Power, Alex Ray, Alex Warstadt, Alexander~W. Kocurek, Ali Safaya, Ali Tazarv, Alice Xiang, Alicia Parrish, Allen Nie, Aman Hussain, Amanda Askell, Amanda Dsouza, Ambrose Slone, Ameet Rahane, Anantharaman~S. Iyer, Anders~Johan Andreassen, Andrea Madotto, Andrea Santilli, Andreas Stuhlm{\"u}ller, Andrew~M. Dai, Andrew La, Andrew~Kyle Lampinen, Andy Zou, Angela Jiang, Angelica Chen, Anh Vuong, Animesh Gupta, Anna Gottardi, Antonio Norelli, Anu Venkatesh, Arash Gholamidavoodi, Arfa Tabassum, Arul Menezes, Arun Kirubarajan, Asher Mullokandov, Ashish Sabharwal, Austin Herrick, Avia Efrat, Aykut Erdem, Ayla Karaka{\c{s}}, B.~Ryan Roberts, Bao~Sheng Loe, Barret Zoph, Bart{\l}omiej Bojanowski, Batuhan {\"O}zyurt, Behnam Hedayatnia, Behnam Neyshabur, Benjamin Inden, Benno Stein, Berk Ekmekci,
  Bill~Yuchen Lin, Blake Howald, Bryan Orinion, Cameron Diao, Cameron Dour, Catherine Stinson, Cedrick Argueta, Cesar Ferri, Chandan Singh, Charles Rathkopf, Chenlin Meng, Chitta Baral, Chiyu Wu, Chris Callison-Burch, Christopher Waites, Christian Voigt, Christopher~D Manning, Christopher Potts, Cindy Ramirez, Clara~E. Rivera, Clemencia Siro, Colin Raffel, Courtney Ashcraft, Cristina Garbacea, Damien Sileo, Dan Garrette, Dan Hendrycks, Dan Kilman, Dan Roth, C.~Daniel Freeman, Daniel Khashabi, Daniel Levy, Daniel~Mosegu{\'\i} Gonz{\'a}lez, Danielle Perszyk, Danny Hernandez, Danqi Chen, Daphne Ippolito, Dar Gilboa, David Dohan, David Drakard, David Jurgens, Debajyoti Datta, Deep Ganguli, Denis Emelin, Denis Kleyko, Deniz Yuret, Derek Chen, Derek Tam, Dieuwke Hupkes, Diganta Misra, Dilyar Buzan, Dimitri~Coelho Mollo, Diyi Yang, Dong-Ho Lee, Dylan Schrader, Ekaterina Shutova, Ekin~Dogus Cubuk, Elad Segal, Eleanor Hagerman, Elizabeth Barnes, Elizabeth Donoway, Ellie Pavlick, Emanuele Rodol{\`a}, Emma Lam, Eric
  Chu, Eric Tang, Erkut Erdem, Ernie Chang, Ethan~A Chi, Ethan Dyer, Ethan Jerzak, Ethan Kim, Eunice~Engefu Manyasi, Evgenii Zheltonozhskii, Fanyue Xia, Fatemeh Siar, Fernando Mart{\'\i}nez-Plumed, Francesca Happ{\'e}, Francois Chollet, Frieda Rong, Gaurav Mishra, Genta~Indra Winata, Gerard de~Melo, Germ{\`a}n Kruszewski, Giambattista Parascandolo, Giorgio Mariani, Gloria~Xinyue Wang, Gonzalo Jaimovitch-Lopez, Gregor Betz, Guy Gur-Ari, Hana Galijasevic, Hannah Kim, Hannah Rashkin, Hannaneh Hajishirzi, Harsh Mehta, Hayden Bogar, Henry Francis~Anthony Shevlin, Hinrich Schuetze, Hiromu Yakura, Hongming Zhang, Hugh~Mee Wong, Ian Ng, Isaac Noble, Jaap Jumelet, Jack Geissinger, Jackson Kernion, Jacob Hilton, Jaehoon Lee, Jaime~Fern{\'a}ndez Fisac, James~B Simon, James Koppel, James Zheng, James Zou, Jan Kocon, Jana Thompson, Janelle Wingfield, Jared Kaplan, Jarema Radom, Jascha Sohl-Dickstein, Jason Phang, Jason Wei, Jason Yosinski, Jekaterina Novikova, Jelle Bosscher, Jennifer Marsh, Jeremy Kim, Jeroen Taal, Jesse
  Engel, Jesujoba Alabi, Jiacheng Xu, Jiaming Song, Jillian Tang, Joan Waweru, John Burden, John Miller, John~U. Balis, Jonathan Batchelder, Jonathan Berant, J{\"o}rg Frohberg, Jos Rozen, Jose Hernandez-Orallo, Joseph Boudeman, Joseph Guerr, Joseph Jones, Joshua~B. Tenenbaum, Joshua~S. Rule, Joyce Chua, Kamil Kanclerz, Karen Livescu, Karl Krauth, Karthik Gopalakrishnan, Katerina Ignatyeva, Katja Markert, Kaustubh Dhole, Kevin Gimpel, Kevin Omondi, Kory~Wallace Mathewson, Kristen Chiafullo, Ksenia Shkaruta, Kumar Shridhar, Kyle McDonell, Kyle Richardson, Laria Reynolds, Leo Gao, Li~Zhang, Liam Dugan, Lianhui Qin, Lidia Contreras-Ochando, Louis-Philippe Morency, Luca Moschella, Lucas Lam, Lucy Noble, Ludwig Schmidt, Luheng He, Luis Oliveros-Col{\'o}n, Luke Metz, L{\"u}tfi~Kerem Senel, Maarten Bosma, Maarten Sap, Maartje~Ter Hoeve, Maheen Farooqi, Manaal Faruqui, Mantas Mazeika, Marco Baturan, Marco Marelli, Marco Maru, Maria~Jose Ramirez-Quintana, Marie Tolkiehn, Mario Giulianelli, Martha Lewis, Martin
  Potthast, Matthew~L Leavitt, Matthias Hagen, M{\'a}ty{\'a}s Schubert, Medina~Orduna Baitemirova, Melody Arnaud, Melvin McElrath, Michael~Andrew Yee, Michael Cohen, Michael Gu, Michael Ivanitskiy, Michael Starritt, Michael Strube, Micha{\l} Sw{\k{e}}drowski, Michele Bevilacqua, Michihiro Yasunaga, Mihir Kale, Mike Cain, Mimee Xu, Mirac Suzgun, Mitch Walker, Mo~Tiwari, Mohit Bansal, Moin Aminnaseri, Mor Geva, Mozhdeh Gheini, Mukund~Varma T, Nanyun Peng, Nathan~Andrew Chi, Nayeon Lee, Neta Gur-Ari Krakover, Nicholas Cameron, Nicholas Roberts, Nick Doiron, Nicole Martinez, Nikita Nangia, Niklas Deckers, Niklas Muennighoff, Nitish~Shirish Keskar, Niveditha~S. Iyer, Noah Constant, Noah Fiedel, Nuan Wen, Oliver Zhang, Omar Agha, Omar Elbaghdadi, Omer Levy, Owain Evans, Pablo Antonio~Moreno Casares, Parth Doshi, Pascale Fung, Paul~Pu Liang, Paul Vicol, Pegah Alipoormolabashi, Peiyuan Liao, Percy Liang, Peter~W Chang, Peter Eckersley, Phu~Mon Htut, Pinyu Hwang, Piotr Mi{\l}kowski, Piyush Patil, Pouya Pezeshkpour,
  Priti Oli, Qiaozhu Mei, Qing Lyu, Qinlang Chen, Rabin Banjade, Rachel~Etta Rudolph, Raefer Gabriel, Rahel Habacker, Ramon Risco, Rapha{\"e}l Milli{\`e}re, Rhythm Garg, Richard Barnes, Rif~A. Saurous, Riku Arakawa, Robbe Raymaekers, Robert Frank, Rohan Sikand, Roman Novak, Roman Sitelew, Ronan~Le Bras, Rosanne Liu, Rowan Jacobs, Rui Zhang, Russ Salakhutdinov, Ryan~Andrew Chi, Seungjae~Ryan Lee, Ryan Stovall, Ryan Teehan, Rylan Yang, Sahib Singh, Saif~M. Mohammad, Sajant Anand, Sam Dillavou, Sam Shleifer, Sam Wiseman, Samuel Gruetter, Samuel~R. Bowman, Samuel~Stern Schoenholz, Sanghyun Han, Sanjeev Kwatra, Sarah~A. Rous, Sarik Ghazarian, Sayan Ghosh, Sean Casey, Sebastian Bischoff, Sebastian Gehrmann, Sebastian Schuster, Sepideh Sadeghi, Shadi Hamdan, Sharon Zhou, Shashank Srivastava, Sherry Shi, Shikhar Singh, Shima Asaadi, Shixiang~Shane Gu, Shubh Pachchigar, Shubham Toshniwal, Shyam Upadhyay, Shyamolima~Shammie Debnath, Siamak Shakeri, Simon Thormeyer, Simone Melzi, Siva Reddy, Sneha~Priscilla Makini,
  Soo-Hwan Lee, Spencer Torene, Sriharsha Hatwar, Stanislas Dehaene, Stefan Divic, Stefano Ermon, Stella Biderman, Stephanie Lin, Stephen Prasad, Steven Piantadosi, Stuart Shieber, Summer Misherghi, Svetlana Kiritchenko, Swaroop Mishra, Tal Linzen, Tal Schuster, Tao Li, Tao Yu, Tariq Ali, Tatsunori Hashimoto, Te-Lin Wu, Th{\'e}o Desbordes, Theodore Rothschild, Thomas Phan, Tianle Wang, Tiberius Nkinyili, Timo Schick, Timofei Kornev, Titus Tunduny, Tobias Gerstenberg, Trenton Chang, Trishala Neeraj, Tushar Khot, Tyler Shultz, Uri Shaham, Vedant Misra, Vera Demberg, Victoria Nyamai, Vikas Raunak, Vinay~Venkatesh Ramasesh, vinay~uday prabhu, Vishakh Padmakumar, Vivek Srikumar, William Fedus, William Saunders, William Zhang, Wout Vossen, Xiang Ren, Xiaoyu Tong, Xinran Zhao, Xinyi Wu, Xudong Shen, Yadollah Yaghoobzadeh, Yair Lakretz, Yangqiu Song, Yasaman Bahri, Yejin Choi, Yichi Yang, Sophie Hao, Yifu Chen, Yonatan Belinkov, Yu~Hou, Yufang Hou, Yuntao Bai, Zachary Seid, Zhuoye Zhao, Zijian Wang, Zijie~J. Wang,
  Zirui Wang, and Ziyi Wu. 2023.
\newblock \href {https://openreview.net/forum?id=uyTL5Bvosj} {Beyond the imitation game: {Q}uantifying and extrapolating the capabilities of language models}.
\newblock \emph{Transactions on Machine Learning Research}.
\newblock Featured Certification.

\bibitem[{Tigges et~al.(2024)Tigges, Hanna, Yu, and Biderman}]{tigges2024llm}
Curt Tigges, Michael Hanna, Qinan Yu, and Stella Biderman. 2024.
\newblock \href {https://proceedings.neurips.cc/paper\%5Ffiles/paper/2024/file/47c7edadfee365b394b2a3bd416048da-Paper-Conference.pdf} {{LLM} circuit analyses are consistent across training and scale}.
\newblock In \emph{Advances in Neural Information Processing Systems}, volume~37, pages 40699--40731. Curran Associates, Inc.

\bibitem[{Touvron et~al.(2023)Touvron, Martin, Stone, Albert, Almahairi, Babaei, Bashlykov, Batra, Bhargava, Bhosale, Bikel, Blecher, Ferrer, Chen, Cucurull, Esiobu, Fernandes, Fu, Fu, Fuller, Gao, Goswami, Goyal, Hartshorn, Hosseini, Hou, Inan, Kardas, Kerkez, Khabsa, Kloumann, Korenev, Koura, Lachaux, Lavril, Lee, Liskovich, Lu, Mao, Martinet, Mihaylov, Mishra, Molybog, Nie, Poulton, Reizenstein, Rungta, Saladi, Schelten, Silva, Smith, Subramanian, Tan, Tang, Taylor, Williams, Kuan, Xu, Yan, Zarov, Zhang, Fan, Kambadur, Narang, Rodriguez, Stojnic, Edunov, and Scialom}]{touvron2023llama2}
Hugo Touvron, Louis Martin, Kevin Stone, Peter Albert, Amjad Almahairi, Yasmine Babaei, Nikolay Bashlykov, Soumya Batra, Prajjwal Bhargava, Shruti Bhosale, Dan Bikel, Lukas Blecher, Cristian~Canton Ferrer, Moya Chen, Guillem Cucurull, David Esiobu, Jude Fernandes, Jeremy Fu, Wenyin Fu, Brian Fuller, Cynthia Gao, Vedanuj Goswami, Naman Goyal, Anthony Hartshorn, Saghar Hosseini, Rui Hou, Hakan Inan, Marcin Kardas, Viktor Kerkez, Madian Khabsa, Isabel Kloumann, Artem Korenev, Punit~Singh Koura, Marie-Anne Lachaux, Thibaut Lavril, Jenya Lee, Diana Liskovich, Yinghai Lu, Yuning Mao, Xavier Martinet, Todor Mihaylov, Pushkar Mishra, Igor Molybog, Yixin Nie, Andrew Poulton, Jeremy Reizenstein, Rashi Rungta, Kalyan Saladi, Alan Schelten, Ruan Silva, Eric~Michael Smith, Ranjan Subramanian, Xiaoqing~Ellen Tan, Binh Tang, Ross Taylor, Adina Williams, Jian~Xiang Kuan, Puxin Xu, Zheng Yan, Iliyan Zarov, Yuchen Zhang, Angela Fan, Melanie Kambadur, Sharan Narang, Aurelien Rodriguez, Robert Stojnic, Sergey Edunov, and Thomas
  Scialom. 2023.
\newblock \href {https://arxiv.org/abs/2307.09288} {Llama 2: {O}pen foundation and fine-tuned chat models}.
\newblock \emph{ArXiv preprint 2307.09288}.

\bibitem[{Van~der Wal et~al.(2024)Van~der Wal, Bachmann, Leidinger, van Maanen, Zuidema, and Schulz}]{van2024undesirable}
Oskar Van~der Wal, Dominik Bachmann, Alina Leidinger, Leendert van Maanen, Willem Zuidema, and Katrin Schulz. 2024.
\newblock \href {https://jair.org/index.php/jair/article/view/15195} {Undesirable biases in {NLP}: {A}ddressing challenges of measurement}.
\newblock \emph{Journal of Artificial Intelligence Research}, 79:1--40.

\bibitem[{Van~der Wal et~al.(2022)Van~der Wal, Jumelet, Schulz, and Zuidema}]{wal2022birth}
Oskar Van~der Wal, Jaap Jumelet, Katrin Schulz, and Willem Zuidema. 2022.
\newblock \href {https://doi.org/10.18653/v1/2022.gebnlp-1.8} {The birth of bias: {A} case study on the evolution of gender bias in an {E}nglish language model}.
\newblock In \emph{Proceedings of the 4th Workshop on Gender Bias in Natural Language Processing (GeBNLP)}, pages 75--75, Seattle, Washington. Association for Computational Linguistics.

\bibitem[{Voita and Titov(2020)}]{voita-titov-2020-information}
Elena Voita and Ivan Titov. 2020.
\newblock \href {https://doi.org/10.18653/v1/2020.emnlp-main.14} {Information-theoretic probing with minimum description length}.
\newblock In \emph{Proceedings of the 2020 Conference on Empirical Methods in Natural Language Processing (EMNLP)}, pages 183--196, Online. Association for Computational Linguistics.

\bibitem[{Wang et~al.(2023)Wang, Variengien, Conmy, Shlegeris, and Steinhardt}]{wang2022interpretability}
Kevin~Ro Wang, Alexandre Variengien, Arthur Conmy, Buck Shlegeris, and Jacob Steinhardt. 2023.
\newblock \href {https://openreview.net/pdf?id=NpsVSN6o4ul} {Interpretability in the wild: {A} circuit for indirect object identification in {GPT-2} small}.
\newblock In \emph{The Eleventh International Conference on Learning Representations, {ICLR} 2023, Kigali, Rwanda, May 1-5, 2023}.

\bibitem[{Warstadt et~al.(2020)Warstadt, Parrish, Liu, Mohananey, Peng, Wang, and Bowman}]{warstadt-etal-2020-blimp-benchmark}
Alex Warstadt, Alicia Parrish, Haokun Liu, Anhad Mohananey, Wei Peng, Sheng-Fu Wang, and Samuel~R. Bowman. 2020.
\newblock \href {https://doi.org/10.1162/tacl_a_00321} {{BLiMP}: {T}he benchmark of linguistic minimal pairs for {E}nglish}.
\newblock \emph{Transactions of the Association for Computational Linguistics}, 8:377--392.

\bibitem[{Welbl et~al.(2017)Welbl, Liu, and Gardner}]{welbl-etal-2017-crowdsourcing}
Johannes Welbl, Nelson~F. Liu, and Matt Gardner. 2017.
\newblock \href {https://doi.org/10.18653/v1/W17-4413} {Crowdsourcing multiple choice science questions}.
\newblock In \emph{Proceedings of the 3rd Workshop on Noisy User-generated Text}, pages 94--106, Copenhagen, Denmark. Association for Computational Linguistics.

\bibitem[{Zeng et~al.(2023)Zeng, Liu, Du, Wang, Lai, Ding, Yang, Xu, Zheng, Xia, Tam, Ma, Xue, Zhai, Chen, Liu, Zhang, Dong, and Tang}]{zeng2022glm}
Aohan Zeng, Xiao Liu, Zhengxiao Du, Zihan Wang, Hanyu Lai, Ming Ding, Zhuoyi Yang, Yifan Xu, Wendi Zheng, Xiao Xia, Weng~Lam Tam, Zixuan Ma, Yufei Xue, Jidong Zhai, Wenguang Chen, Zhiyuan Liu, Peng Zhang, Yuxiao Dong, and Jie Tang. 2023.
\newblock \href {https://openreview.net/pdf?id=-Aw0rrrPUF} {{GLM-130B}: {A}n open bilingual pre-trained model}.
\newblock In \emph{The Eleventh International Conference on Learning Representations, {ICLR} 2023, Kigali, Rwanda, May 1-5, 2023}.

\bibitem[{Zhai et~al.(2023)Zhai, Likhomanenko, Littwin, Busbridge, Ramapuram, Zhang, Gu, and Susskind}]{zhai2023stabilizing}
Shuangfei Zhai, Tatiana Likhomanenko, Etai Littwin, Dan Busbridge, Jason Ramapuram, Yizhe Zhang, Jiatao Gu, and Joshua~M. Susskind. 2023.
\newblock \href {https://proceedings.mlr.press/v202/zhai23a.html} {Stabilizing transformer training by preventing attention entropy collapse}.
\newblock In \emph{International Conference on Machine Learning, {ICML} 2023, 23-29 July 2023, Honolulu, Hawaii, {USA}}, volume 202 of \emph{Proceedings of Machine Learning Research}, pages 40770--40803. {PMLR}.

\bibitem[{Zoph et~al.(2022)Zoph, Bello, Kumar, Du, Huang, Dean, Shazeer, and Fedus}]{zoph2022st}
Barret Zoph, Irwan Bello, Sameer Kumar, Nan Du, Yanping Huang, Jeff Dean, Noam Shazeer, and William Fedus. 2022.
\newblock \href {https://arxiv.org/abs/2202.08906} {{ST-MoE}: {D}esigning stable and transferable sparse expert models}.
\newblock \emph{ArXiv preprint 2202.08906}.

\end{thebibliography}
